\documentclass[AMS,STIX1COL]{WileyNJD-v2}

\usepackage{graphicx}
\usepackage{hyperref}
\hypersetup{
    colorlinks=true,
    linkcolor=blue,
    filecolor=magenta,      
    urlcolor=cyan,
}
\usepackage{subcaption}

\articletype{Article Type}%

\received{26 April 2016}
\revised{6 June 2016}
\accepted{6 June 2016}

\newcommand{\mbold}[1]{\mbox{\boldmath ${#1}$}}
\graphicspath{{./figures/}}
\begin{document}

\title{Random Forest (RF) Kernel for Regression, Classification and Survival \protect\thanks{This is an example for title footnote.}}

\author[1]{Dai Feng*}

\author[2]{Richard Baumgartner}

\authormark{AUTHOR ONE \textsc{et al}}

\address[1]{\orgdiv{Data and Statistical Sciences}, \orgname{AbbVie Inc.}, \orgaddress{\state{North Chicago, Illinois}, \country{United States of America}}}

\address[2]{\orgdiv{Merck \& Co., Inc.}, \orgname{Kenilworth, NJ}, \country{United States of America}}

\corres{*Dai Feng, Data and Statistical Sciences, AbbVie Inc. \email{dai.feng@abbvie.com}}

\presentaddress{This is sample for present address text this is sample for present address text}

\abstract[Summary]{Breiman’s random forest (RF) can be interpreted as an implicit kernel generator, where the ensuing proximity matrix represents the data-driven RF kernel. Kernel perspective on the RF has been used to develop a principled framework for theoretical investigation of its statistical properties. However, practical utility of the links between kernels and the RF has not been widely explored and systematically evaluated. 

Focus of our work is investigation of the interplay between kernel methods and the RF. We elucidate the performance and properties of the data driven RF kernels used by regularized linear models in a comprehensive simulation study comprising of continuous, binary and survival targets. We show that for continuous and survival targets, the RF kernels are competitive to RF in higher dimensional scenarios with larger number of noisy features. For the binary target, the RF kernel and RF exhibit comparable performance. As the RF kernel asymptotically converges to the Laplace kernel, we included it in our evaluation. For most simulation setups, the RF and RF kernel outperformed the Laplace kernel. Nevertheless, in some cases the Laplace kernel was competitive, showing its potential value for applications. We also provide the results from real life data sets for the regression, classification and survival to illustrate how these insights may be leveraged in practice. 

Finally, we discuss further extensions of the RF kernels in the context of interpretable prototype and landmarking classification, regression and survival. We outline future line of research for kernels furnished by Bayesian counterparts of the RF.
}


\jnlcitation{\cname{%
\author{Dai Feng}, 
\author{Richard Baumgartner}(\cyear{2020})}, 
\ctitle{Random Forest (RF) Kernel for Regression, Classification and Survival} 
}

\maketitle


\section{Introduction}\label{sec1}
Random forest (RF) has been a successful and time-proven statistical machine learning method \cite{biau2016}. At first, RF was developed for classification and regression \cite{breiman2000}. Recently, it has been extended and adopted for additional types of targets such as time-to-event or ordered outcomes \cite{ishwaran2019}. RF belongs to the ensemble methods, where “base” tree learners are grown on bootstrapped samples of the training data set and then their predictions are aggregated to yield a final prediction. RF was conceived originally under the frequentist framework. However, Bayesian counterparts e.g. Mondrian random forest were also proposed \cite{balog2016}. 

In Ref \cite{breiman2000}, Breiman pointed out an alternative interpretation of the RF as a kernel generator.  The $n \times n$ proximity matrix (where $n$ is the number of samples) naturally ensuing from the construction of the RF plays here a key role. Each entry of the RF proximity matrix is an estimate of the probability that two points end up in the same terminal node \cite{breiman2000}. It is a symmetric positive-semidefinite matrix and it can be interpreted as a kernel akin to those previously proposed for the kernel methods \cite{breiman2000},\cite{scornet2016}.  To note, in Bayesian framework, Mondrian kernel denotes also the empirical frequency with which two points end up in the same partition cell of a Mondrian sample \cite{balog2016}. Moreover the examples of frequently used (analytical) kernels include linear kernel, radial basis function (RBF), polynomial kernels, etc. \cite{herbich2001},\cite{schoelkopf2001}.  Asymptotically the RF kernel converges to the Laplace kernel \cite{breiman2000}. Similar convergence results were obtained also for Mondrian forests and Bayesian additive regression trees (BART) \cite{balog2016, linero2017}.

Supervised kernel methods as usually applied, fit linear models in non-linear feature spaces that are induced by the kernels.  Popular choices in this class of algorithms include support vector machines (SVMs) and kernel ridge regression \cite{herbich2001},\cite{schoelkopf2001} or their refinements e.g. generalized "kernel" elastic net \cite{sokolov2016}. Bayesian approach to the kernel methods is represented by the Gaussian processes \cite{davies2014},\cite{rasmussen2006}. 

Relevant to our work is also similarity/dissimilarity based learning \cite{chen2009},\cite{pekalska2001},\cite{balcan2008} that was proposed for classification and regression. In similarity/dissimilarity learning the kernel entries are explicitly interpreted as pairwise similarities/dissimilarities between points (samples). RF proximity matrix or RF kernel fits readily into this paradigm.

The kernel interpretation of the RF was further explored and expounded theoretically to investigate its statistical properties such as asymptotic convergence of RF and RF kernel estimates \cite{scornet2016}. On the other hand, there has been also interest in the use of algorithms based on the RF kernel \cite{davies2014} in practice.  In the Ref \cite{davies2014} performance of RF kernels was found competitive for regression tasks on various data sets from the UCI repository. 

In our work we are building on the  \cite{davies2014}. Our focus is investigation of the RF kernel based algorithms in regression, classification and survival (time-to-event outcomes) and elucidation of their performance characteristics. The manuscript is organized as follows: Section 2 introduces the theoretical framework of the RF kernel for targets of interest, Section 3 provides a motivational example using the well known Fisher Iris data, Section 4 details a simulation study that systematically evaluates performance of the RF kernel in various scenarios, Section 5 summarizes the results on real life data sets and Section 6 provides discussion, conclusions and future research directions.

\section{Methodology}\label{sec2}

\subsection{Terminology}\label{sec2sub2}
Following Breiman \cite{breiman2000} and Refs. \cite{ishwaran2019} and \cite{scornet2016} we consider a supervised learning problem, where training set $D_n=\{(\mbold{X_1},Y_1), (\mbold{X_2},Y_2),\ldots,({\mbold{X_1},Y_n})\}$ is provided. $\mbold{X_i} \in R^p$ and $Y_i$ can be continuous, binary or survival target. For continuous and binary targets the $Y_i \in R$ and $Y_i\in \{0,1\}$, respectively. The survival targets are assumed to be potentially right censored, i.e. they comprise of a continuous target $Y_i$ or lower bound $C_i$ with the indicator of right censoring $I(C_i<Y_i)$ at $C_i$. We define the vector of targets as $\mbold{Y} = (Y_1,Y_2,\ldots,Y_n)^T$. The goal here is to learn a predictor from $D_n$ to facilitate predictions of the $Y_i$-s on an independent test set, for which only $\mbold{X}$-s are provided.

\subsection{Kernels for Regression, 
 Classification and Survival}\label{sec2sub3}
Kernel methods in the machine learning literature are a class of methods that are formulated in terms a similarity (Gram) matrix $\mbold{K}$. The similarity matrix $K_{i,j}=k(\mbold{X_1},\mbold{X_2})$ represents the similarity between two points $\mbold{X_1}$ and $\mbold{X_2}$.
Kernel methods have been well developed and there is a large body of references covering their different aspects \cite{herbich2001},\cite{schoelkopf2001},\cite{friedmanHastieTibshirani2009}. 
In our work we used two common kernel algorithms, namely kernel Ridge Regression (KRR) and Support Vector Machines (SVMs) for regression and classification and survival, respectively. For the two class classification, we developed the KRR model with targets of -1 and 1 denoting the two classes. The predicted class label was obtained by thresholding around 0.

KRR is a kernelized version of the traditional linear ridge regression with the L2-norm penalty. Given the kernel matrix $\mbold{K}$ estimated from the training set, first the coefficients $\mbold{\alpha}$ of the (linear) KRR predictor in the non-linear feature space induced by the kernel $k(.,.)$ are 
obtained:
\begin{eqnarray}
\mbold{\alpha}&=&(\mbold{K}+\lambda \mbold{I_n})^{-1}\mbold{Y}
\end{eqnarray}
where $\lambda$ is the regularization parameter.

The KRR predictor $h_{\text{KRR}}(\mbold{X})$ is given as:  

\begin{eqnarray}
h_{KRR}(\mbold{X})&=&\sum_{i=1}^{n} \alpha_i
\mbold{k(X_i,X)}=\mbold{Y^T}(\mbold{K}+\lambda\mbold{I_n})^{-1}\mbold{K_i}
\label{Eq:KRRPredictor}
\end{eqnarray}
where $\mbold{K_i}=(k(X_1,X), \ldots, k(X_n,X))$.

The survival regression (i.e. time-to-event target) is similar to that for continuous target with the additional challenge posed by accommodation of potential right censoring of the outcomes. Hence, the survival outcome consists of the (continuous) survival time to event $Y_i$ and the indicator of (right) censoring $Y_{surv_{i}}=(Y_i,1-\delta_i)$. Where $\delta_i = 1-I(Y_i > C_i)$ is $0$ for right censored outcomes $Y_i$ censored at lower bound $C_i$. To accommodate the right censoring we used the survival support vector machine regression (SSVM) according to \cite{vanBelle2011},\cite{shiwasvami2007}. The goal of the SSVM is to find a linear predictor $h_{\text{SSVM}}(\mbold{X})$ (also referred to as a prognostic index in \cite{vanBelle2011})  that is concordant with the survival:
\begin{eqnarray}
h_{\text{SSVM}}(\mbold{X})&=&\sum_{i=1}^{n}(\alpha_i-\delta_i\alpha_i^{\ast})\mbold{k(X_i,X)}+b
\label{Eq:SSVMPredictor}
\end{eqnarray}

 The coefficients $\alpha_i$ and $\alpha_i^\ast$ are the Lagrange multipliers furnished by the solution of the SSVM dual (quadratic) optimization problem. As it is the case for the classification and regression, analytical kernels e.g. Laplace, radial basis function, etc. and also the RF kernel can be considered. Note that for the non-censored targets the SSVM solution is equivalent to that of standard SVM regression. The formulations of the primal and dual SSVM optimization problems are given in the online Appendix.

\subsection{Random Forest (RF) and RF Kernel}\label{sec2sub4}
 Random Forest (RF) is defined  as an ensemble of tree predictors grown on bootstrapped samples of a training set\cite{breiman2000}. When considering an ensemble of tree predictors $\{h(.,\Theta_m,D_n), m=1,2,\ldots,M\}$, with $\{h(.,\Theta_m,D_n)$ representing a single tree. The $\Theta_1, \Theta_2,\ldots\Theta_M$ are iid random variables that encode the randomization necessary for the tree construction \cite{scornet2016},\cite{ishwaran2019}.

The RF predictor is obtained as:
\begin{eqnarray}
h_{\text{RF}}(\mbold{X},\Theta_1,\ldots,\Theta_m,D_n)&=& \frac{1}{M}\sum_{m=1}^M h(\mbold{X},\Theta_m,D_n)
 \end{eqnarray}
RF kernel ensuing from the RF is defined as a probability that  $\mbold{X_1}$ and $\mbold{X_2}$ are in the same terminal node $R_k(\Theta_m$) \cite{breiman2000},\cite{scornet2016}.
\begin{eqnarray}
k_{RF}(\mbold{X_1},\mbold{X_2})=\frac{1}{M}
\sum_{m=1}^M \sum_{k=1}^T I(\mbold{X_1},\mbold{X_2} \in R_k(\Theta_m))
\end{eqnarray}

\subsection{RF Kernel Predictors for Regression, Classification and Survival}\label{sec2sub5}

RF kernel predictor for regression was obtained by substituting for the RF kernel in Eq.\ref{Eq:KRRPredictor} as:
\begin{eqnarray}
h_{\text{RF-KRR}}(\mbold{X})&=&\sum_{i=1}^{n} \alpha_i
\mbold{k_{RF}(X_i,X)}=\mbold{Y^T}(\mbold{K_{RF}}+\lambda\mbold{I_n})^{-1}\mbold{K_{RFi}}
\label{Eq:KRRrfpredictor}
\end{eqnarray}
The RF kernel predictor for classification was also obtained as that for regression using Eq.\ref{Eq:KRRrfpredictor}, by building a regression model with target classes denoted as $\{-1,1\}$ and a class prediction threshold of $0$.
Similarly, the RF kernel predictor for survival is given according to the Eq.\ref{Eq:SSVMPredictor}:
\begin{eqnarray}
h_{\text{RF-SSVM}}(\mbold{X})&=&\sum_{i=1}^{n}(\alpha_i-\delta_i\alpha_i^{\ast})\mbold{k_{RF}(X_i,X)}+b
\end{eqnarray}

The code for the simulation and real life data analysis was developed in the R programming language \cite{Rcran}. For the continuous and binary targets the ranger \cite{ranger} implementation of RF was used. The regularization parameter $\lambda$ was chosen at minimum value, such that the matrix $\mbold{K}+\lambda \mbold{I_n}$ was invertible.
To calculate the Laplace kernel for regression and classification, we used the 
 package kernlab \cite{kernlab}.
For the survival target, the survival forest function cforest \cite{cforest} was applied in conjunction with the survivalsvm package \cite{survivalsvm}.  The survivalsvm function was modified to handle a customized kernel, in this case the RF and Laplace kernel. 

In the simulations, all algorithms were applied using their default parameters. In order to further elucidate the impact of tree depth on the simulation results we carried out a sensitivity analysis, where we doubled the minimum size of the tree terminal node to generate more shallow trees. The doubled minimum tree node size equaled to 10 and 2 for regression and classification, respectively. The minimum sum of weights in a terminal node was 14 for survival.

\section{Motivating Example}\label{sec3Mot}
\begin{figure}[ht]
\centerline{
\includegraphics[width=0.8\textwidth,height=0.5\textheight]{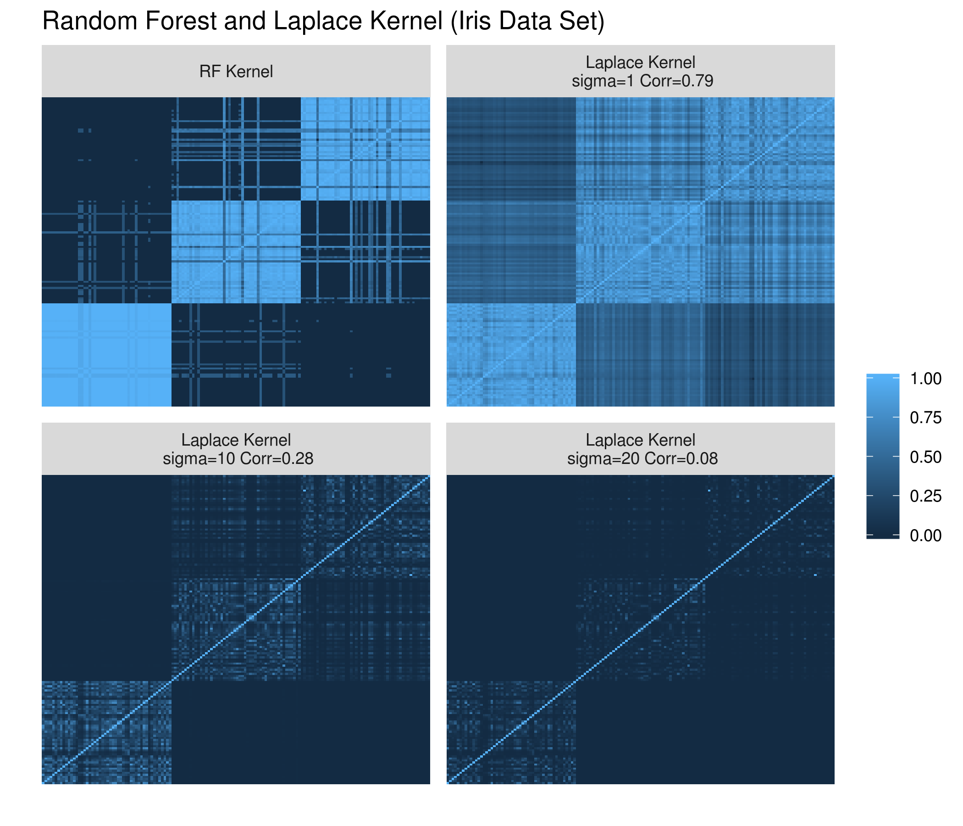}}
\caption{RF Kernel and the Laplace Kernel for the Fisher Iris data set. Corr denotes the matrix correlation given by the Mantel statistic and sigma is the $\sigma$ parameter of the Laplace kernel.\label{fig:iris}}
\end{figure}

As a motivating example we show kernel matrices obtained from the the Fisher's Iris data. The Iris data consists of recordings of three Iris subspecies: Setosa, Versicolor and Virginica (50 samples each recorded
 on 4 numerical features). We compare the RF and Laplace kernels for this data set. 
The Laplace kernel is defined as $k(\mbold{X_1}, \mbold{X_2})=\exp(\frac{-||X_1-X_2||_1}{\sigma}$).
In Figure \ref{fig:iris}, RF kernel is compared with the Laplace kernel for different values of the parameter $\sigma$.

The similarity between RF kernel obtained as a proximity matrix and the Laplace kernel is assessed by the Mantel statistic, i.e. matrix correlation in this case between two similarity matrices \cite{legendre2012}. In the Figure \ref{fig:iris}, the RF kernel captures the underlying structure of the data well and the three classes can be clearly distinguished. Similarly, the Laplace kernels also reflect (with different success) the partitioning of the data in three classes. The Laplace kernel that has the highest Mantel statistics with respect to the RF kernel appears to be the best in terms of the target alignment. 

\begin{figure}
\centerline{\includegraphics[width=0.8\textwidth,height=0.5\textheight]{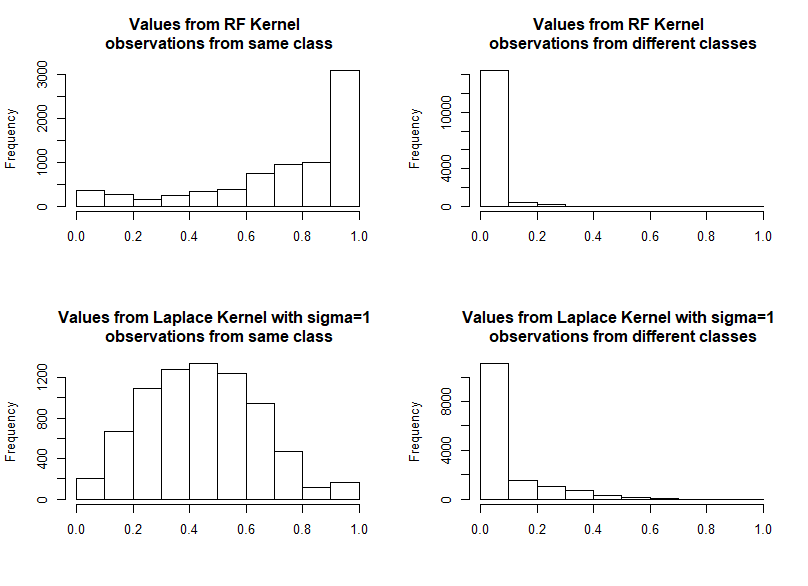}}
\caption{Distributions of values from RF Kernel and the Laplace Kernel with $\sigma=1$ for the Fisher Iris data set.} 
\label{fig:iris-kerVal}
\end{figure}

Furthermore, the RF kernel could characterize a more precise similarity function \cite{balcan2008} leading to more accurate classification results. When RF kernel is compared to the Laplace kernel, the observations from the same class are more similar (closer) to each other than those from different classes. This is demonstrated by the histograms shown in Figure \ref{fig:iris-kerVal}, with the RF kernel histogram peaking at 1 and 0 for the observations from the same class and those from different classes, respectively.

The main goal of this manuscript is to investigate the utility of the RF kernel (proximity) matrix in building predictive models for regression, classification and survival. Due to its asymptotic relationships to the RF kernel \cite{breiman2000}, \cite{balog2016}, Laplace kernel (with $\sigma=1$) was also examined to further elucidate the merits of the RF kernel.

\section{Simulation}\label{sec3}
Simulation scenarios for performance evaluation of RF kernel for regression  were set up according to previously reported simulation benchmarks for continuous targets including
Friedman \cite{friedman}, Meier 1, Meier 2 \cite{meier},  van der Laan \cite{van2007super} and Checkerboard \cite{zhu2015}. These were also adapted for classification and survival.

\subsection{Simulation Setup}\label{sec3sub1}

For each simulation scenario, the predictors were simulated from Uniform (Friedman, Meier 1, Meier 2, van der Laan) or Normal distributions (Checkerboard), respectively. 

Continuous targets were generated as $Y_i=f(\mbold{X_i})+\epsilon_i$. For the definitions of $f(\mbold{X_i})$ for each simulation case see below.

To generate a binary outcome $Y_i$ the continuous outcome was first centered by the median $M$ of its marginal distribution to obtain a balanced two class problem. Then the binary target was generated as a Bernoulli variable with $p_i=prob(Y_i=1|\mbold{X_i})$, where $p_i$ was calculated as follows:
\begin{eqnarray}
p_i &=& \frac{\exp{(f(\mbold{X_i})-M)}}{1+\exp{(f(\mbold{X_i})-M)}}\nonumber
 \end{eqnarray}
where $f(\mbold{X_i})$ is obtained from the continuous models. 

To characterize the intrinsic complexity of the classification problems, we also calculated the Bayes error rate from a large sample of the continuous outcomes ($n=10^{7}$) and subsequently applied the formula $\textnormal{Error}_{\textrm{Bayes}} =1-E\left(\underset{j}{\textrm{max}} Pr(Y=j|\mbold{X})\right), j\in\{0,1\}$ according to \cite{James2013}. The $j\in\{0,1\}$ refer to the class indicators. 

We simulated time-to-event data using Cox proportional hazards model. The survival time $T_i$ for the $i^{th}$ subject was generated with a hazard function given by Eq. (\ref{eqn:hazard}). 
\begin{equation}
\lambda_i(t)=\lambda_0(t)\exp{(f(\mbold{X_i}))}
\label{eqn:hazard}
\end{equation}
Furthermore, we generated censoring times $C_i$ from an exponential distribution controlling the censoring rate. 
Due to censoring, we observe $Y_i = min(T_i, C_i)$ and censoring indicator $\delta_i$.

The five functional relationships $f(\mbold{X_i})$ between the predictors and target for different simulation settings are specified as follows.

1. Friedman. The setup for Friedman was as described in \citet{friedman}.
\begin{eqnarray}
X_{ij} &\sim& Uniform(0, 1), i=1,\ldots,n; j=1,\ldots,p \nonumber\\
\epsilon_i &\sim& N(0,1)\nonumber\\
f(\mbold{X_i}) &=& 10\sin{(\pi X_{i1}X_{i2})} + 20(X_{i3}-0.5)^2+10X_{i4} + 5X_{i5} + \epsilon_i\nonumber
\end{eqnarray}
Bayes error rate for the Friedman classification problem is $0.02$. It is the least complex problem by this measure among those investigated.

2. Checkerboard. In addition to Friedman, we simulated data from a Checkerboard-like model with strong correlation as in Scenario 3 of \citet{zhu2015}.

\begin{eqnarray}
\mbold{X_i} &\sim& N(0, \Sigma_{p\times p}), i=1, \ldots, n \nonumber\\
\epsilon_i &\sim& N(0,1)\nonumber\\
f(\mbold{X_i})&=&2X_{i5}X_{i10}+2X_{i15}X_{i20} + \epsilon_i\nonumber
 \end{eqnarray}
The $(j,k)$ component of $\Sigma$ is equal to $0.9^{|j-k|}$.
Bayes error rate for the Checkerboard classification problem is $0.18$.

3. van der Laan. The setup was studied in van der Laan et al. \cite{van2007super}.
\begin{eqnarray}
X_{ij} &\sim& Uniform(0, 1), i=1,\ldots,n; j=1,\ldots,p \nonumber\\
\epsilon_i &\sim& N(0,0.5)\nonumber\\
f(\mbold{X_i}) &=& \tilde{X}_{i1}\tilde{X}_{i2}+\tilde{X}_{i3}^2+\tilde{X}_{i8}\tilde{X}_{i10}-
\tilde{X}_{i6}^2+ \epsilon_i\nonumber\\
\tilde{X}_i &= &2(X_i-0.5)
\end{eqnarray}
Bayes error rate for the van der Laan classification problem is $0.34$, making it the most complex by this measure among those investigated.

4. Meier 1. This setup was investigated in  Meier et al. \cite{meier}.
\begin{eqnarray}
X_{ij} &\sim& Uniform(0, 1), i=1,\ldots,n; j=1,\ldots,p \nonumber\\
\epsilon_i &\sim& N(0,0.5)\nonumber\\
f(\mbold{X_i}) &=& -\sin(2\tilde{X}_{i1})+\tilde{X}_{i2}^2+\tilde{X}_{i3}-\exp(\tilde{X}_{i4})+ \epsilon_i\nonumber\\
\tilde{X}_i &= &2(X_i-0.5)
\end{eqnarray}
Bayes error rate for the Meier 1 classification problem is $0.28$.

5. Meier 2. This setup was investigated in  Meier et al. \cite{meier} as well.
\begin{eqnarray}
X_{ij} &\sim& Uniform(0, 1), i=1,\ldots,n; j=1,\ldots,p \nonumber\\
\epsilon_i &\sim& N(0,0.5)\nonumber\\
f(\mbold{X_i}) &=& -\tilde{X}_{i1}+(2\tilde{X}_{i2}-1)^2+\frac{\sin(2\pi\tilde{X}_{i3})}
{2-\sin(2\pi\tilde{X}_{i4})}+2\cos(2\pi\tilde{X}_{i4})+4\cos^2(2\pi\tilde{X}_{i4})+ \epsilon_i\nonumber\\
\tilde{X}_i &= &2(X_i-0.5)
\end{eqnarray}
Bayes error  rate for the Meier 2 classification problem is $0.19$.

We used mean squared error (MSE), classification accuracy, and Harrell's concordance index (C-index) \cite{harrell1996} to measure the prediction performance for continuous, binary, and survival data, respectively. For continuous, binary and survival data, we prefer smaller MSE, higher accuracy and larger C-index, respectively. The C-index for survival data is is an estimate of probability of concordance between predicted and observed survival and it is obtained as a ratio of concordant to comparable prognostic index-outcome pairs, respectively \cite{vanBelle2011}. For the definition of the C-index see also the Supplementary information. C-index ranges between 0.5 (random prediction) to 1 (perfectly concordant prognostic index - outcome pairs) \cite{vanBelleJMRL2011}. 

For each functional relationship $f(\mbold{X_i})$ (Friedman, Checkerboard, Meier 1, Meier 2, and van der Laan) and each outcome (continuous, binary, or survival), we simulated data from four scenarios with different samples sizes $n=800$ and $n=1600$ and number of covariates $p=20$ and $p=40$. Within each scenario, we simulated 200 data sets and for each data set we randomly chose 75\% of samples as training data and remaining 25\% as test data.

\subsection{Simulation Results}\label{sec3sub2}
 
\begin{figure}
\begin{subfigure}{.33\textwidth}
  \centering
  \includegraphics[height=0.2\textheight]{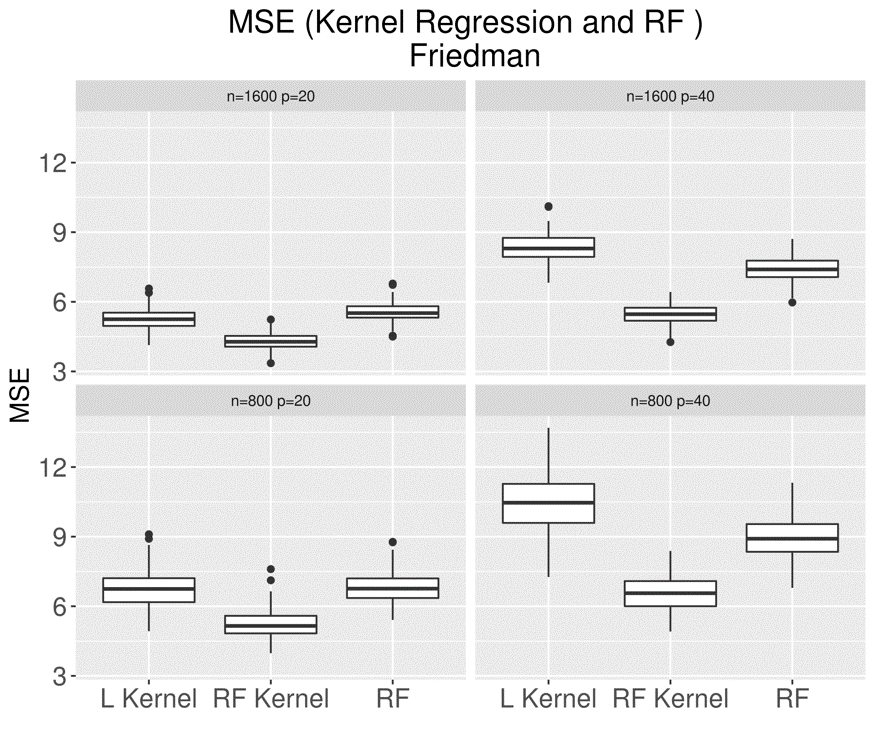}  
  \caption{Continuous MSE}
  \label{fig:sub-first}
\end{subfigure}
\begin{subfigure}{.33\textwidth}
  \centering
  \includegraphics[height=0.2\textheight]{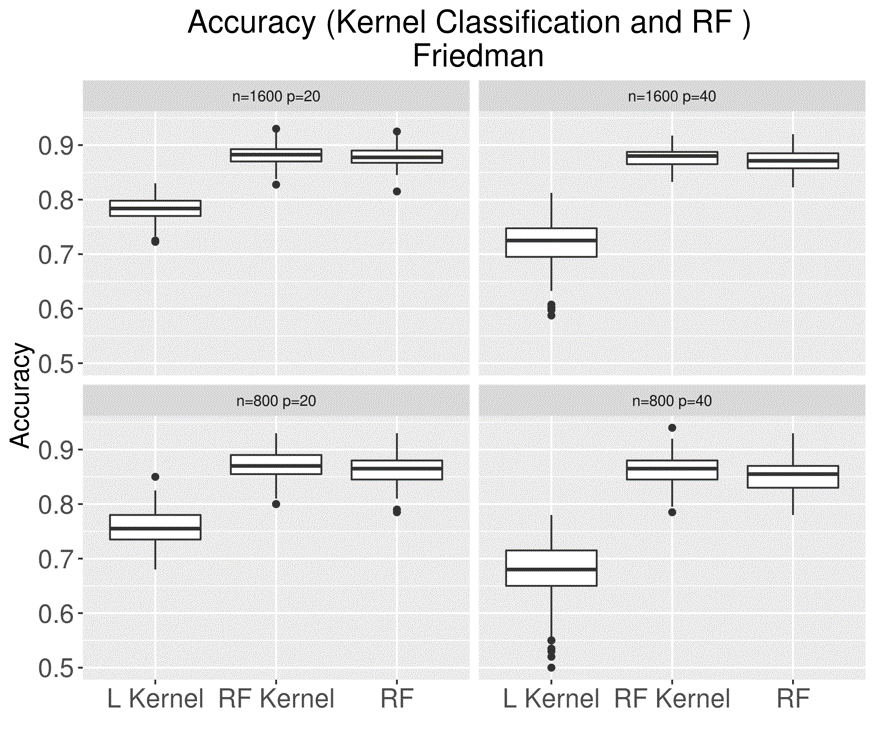}  
  \caption{Binary Accuracy}
  \label{fig:sub-first}
\end{subfigure}
\begin{subfigure}{.33\textwidth}
  \centering
  \includegraphics[height=0.2\textheight, width=0.9\textwidth]{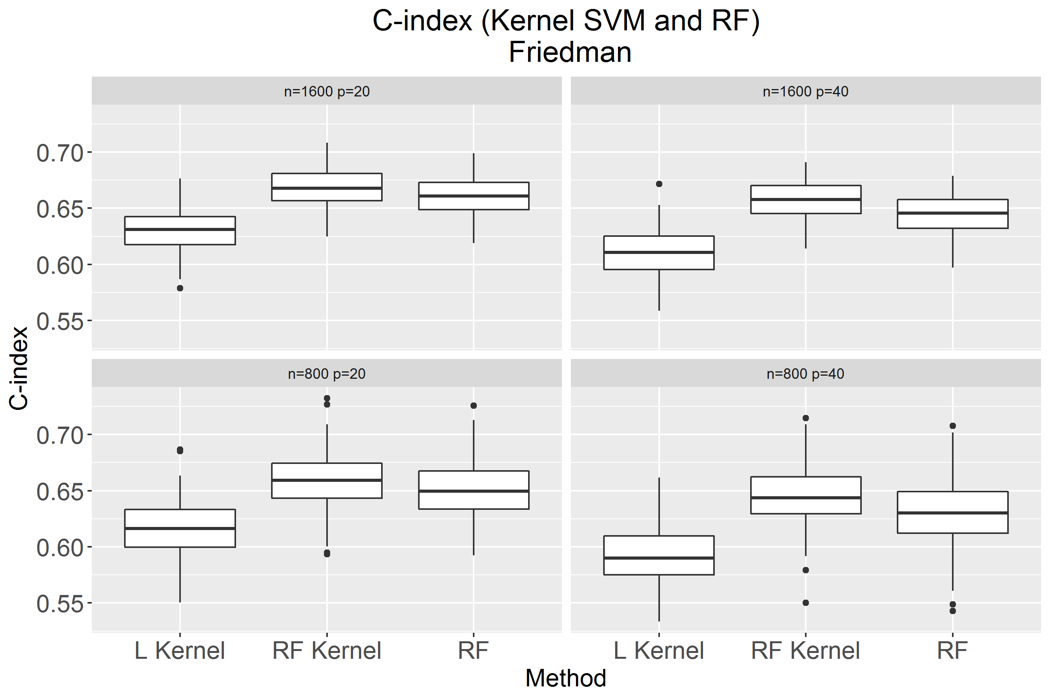}  
  \caption{Survival C-index}
  \label{fig:sub-first}
\end{subfigure}\\
\begin{subfigure}{.33\textwidth}
  \centering
  \includegraphics[height=0.2\textheight]{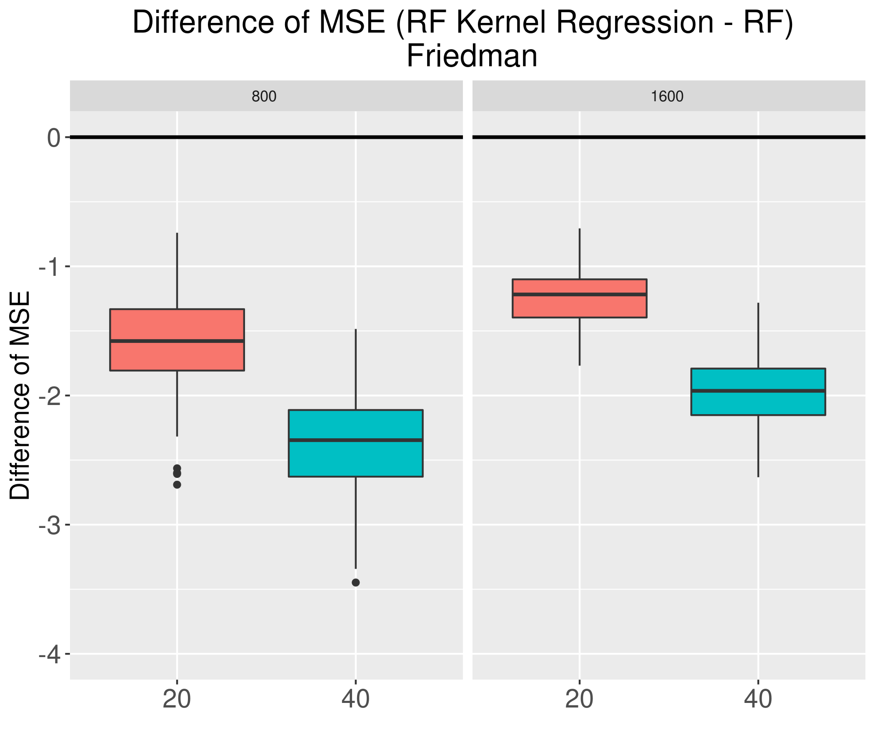}  
  \caption{Continuous Difference of MSE}
  \label{fig:sub-first}
\end{subfigure}
\begin{subfigure}{.33\textwidth}
  \centering
  \includegraphics[height=0.2\textheight]{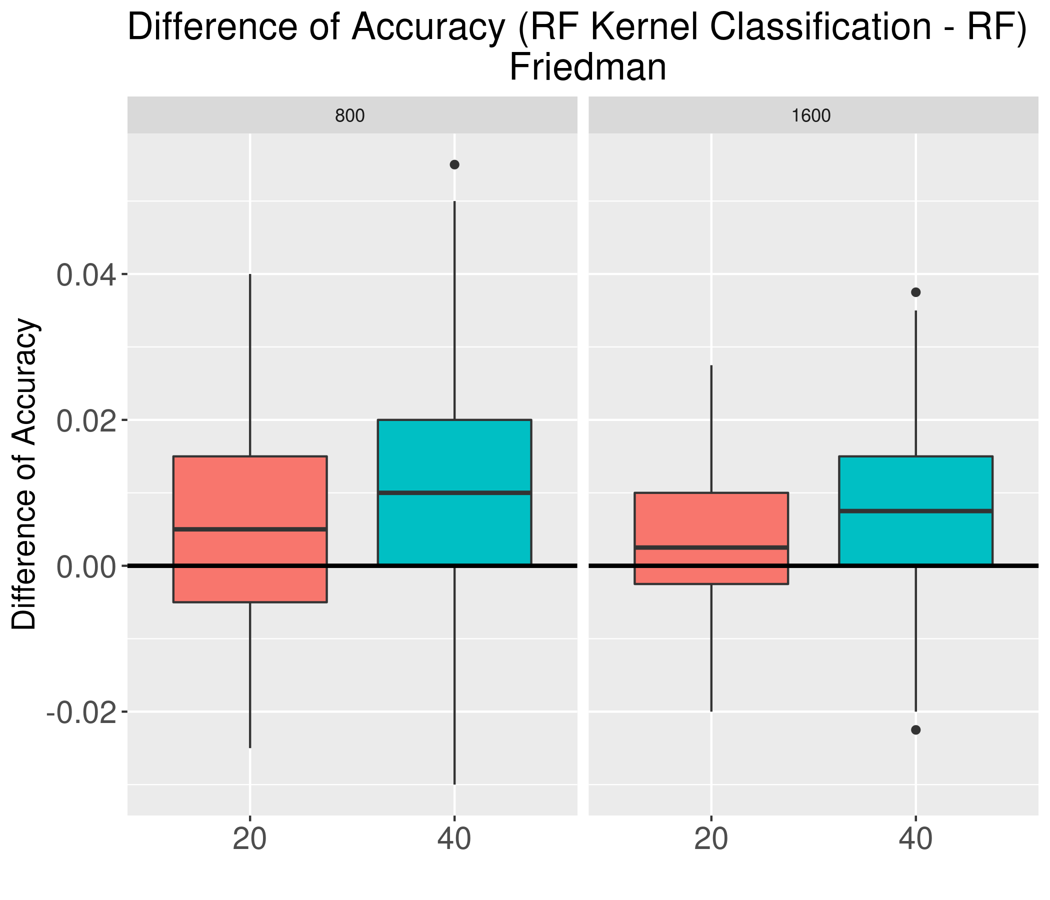} 
  \caption{Binary Difference of Accuracy}
  \label{fig:sub-second}
\end{subfigure}
\begin{subfigure}{.33\textwidth}
  \centering
  \includegraphics[height=0.2\textheight]{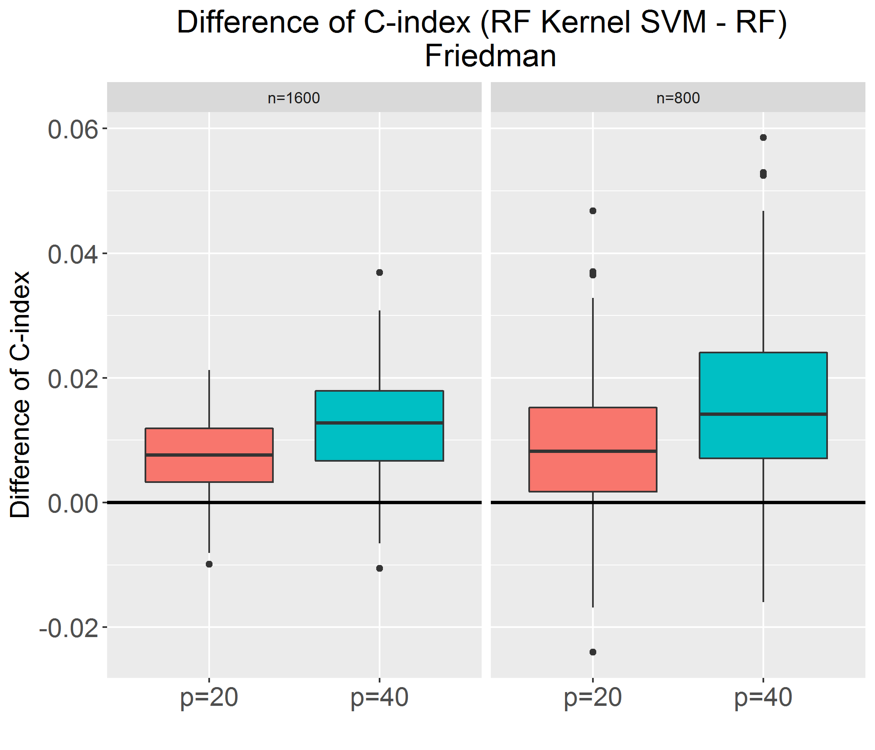} 
  \caption{Survival Difference of C-index}
  \label{fig:sub-second}
\end{subfigure}
\caption{Comparison of MSE, classification accuracy and  C-index for continuous, binary and survival targets, respectively, using RF, RF kernel and Laplace kernel for data simulated from Friedman setting}
\label{fig:Friedman}
\end{figure}

\begin{figure}[ht]
\begin{subfigure}{.33\textwidth}
  \centering
  \includegraphics[height=0.2\textheight]{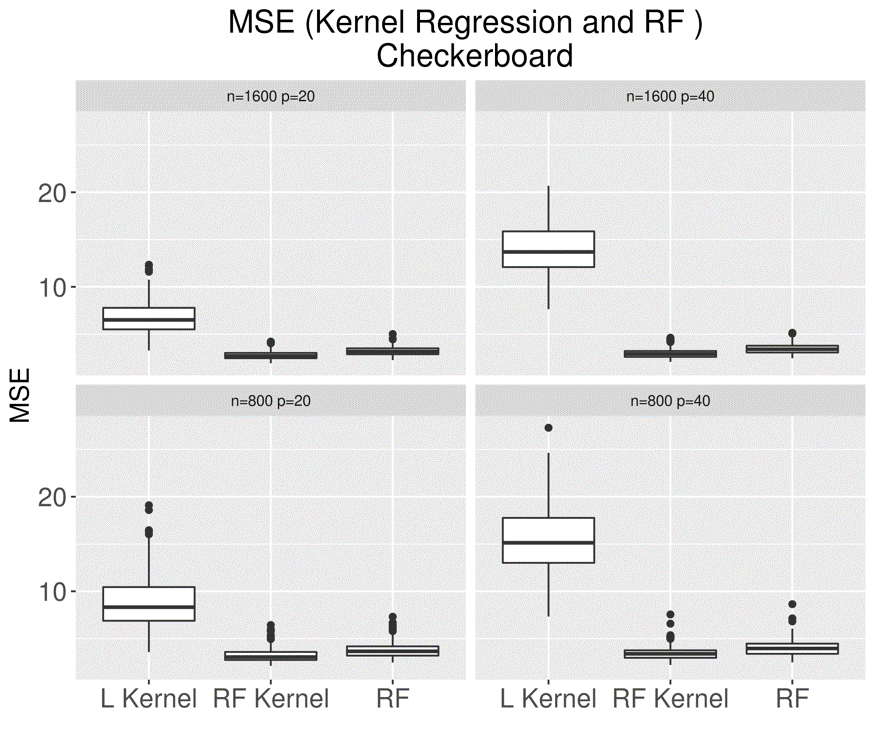}  
  \caption{Continuous MSE}
  \label{fig:sub-first}
\end{subfigure}
\begin{subfigure}{.33\textwidth}
  \centering
  \includegraphics[height=0.2\textheight]{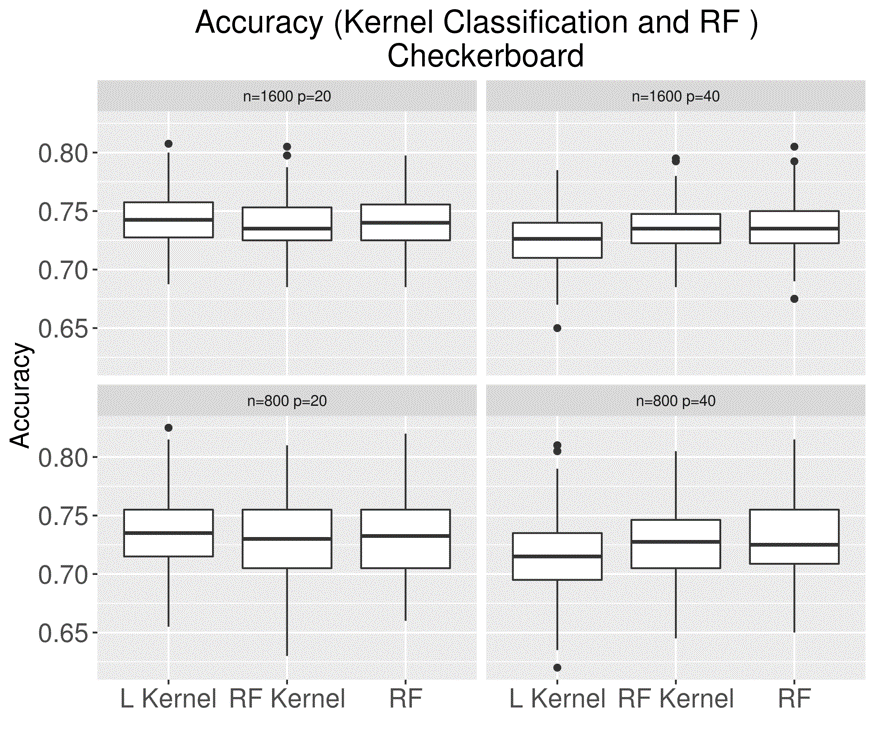}  
  \caption{Binary Accuracy}
  \label{fig:sub-first}
\end{subfigure}
\begin{subfigure}{.33\textwidth}
  \centering
  \includegraphics[height=0.2\textheight, width=0.9\textwidth]{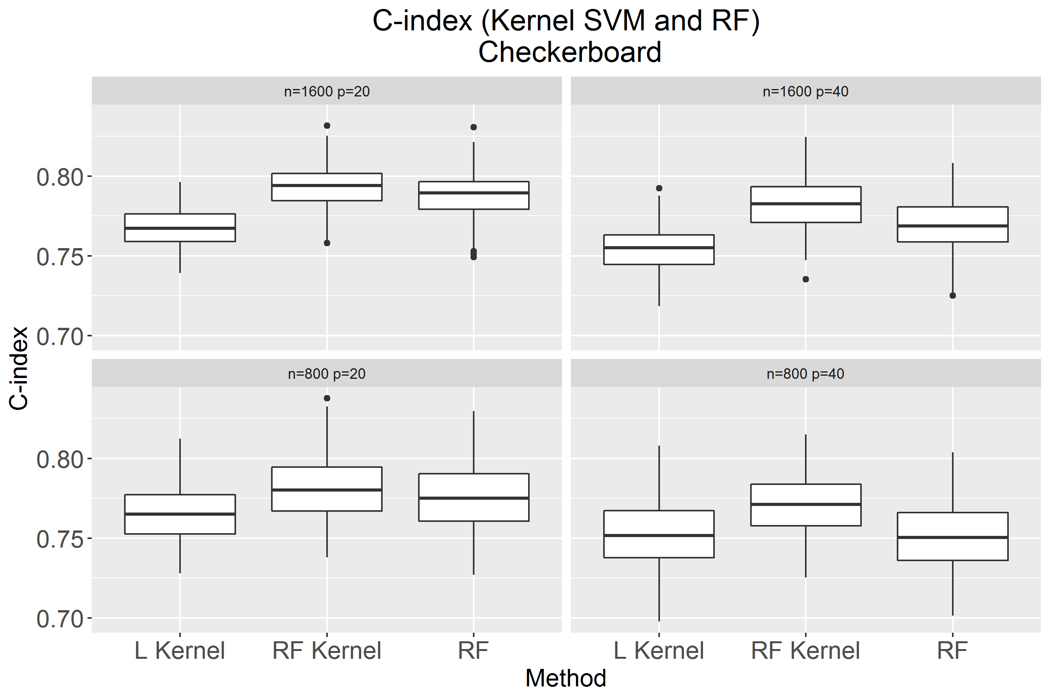}  
  \caption{Survival C-index}
  \label{fig:sub-first}
\end{subfigure}\\
\begin{subfigure}{.33\textwidth}
  \centering
  \includegraphics[height=0.2\textheight]{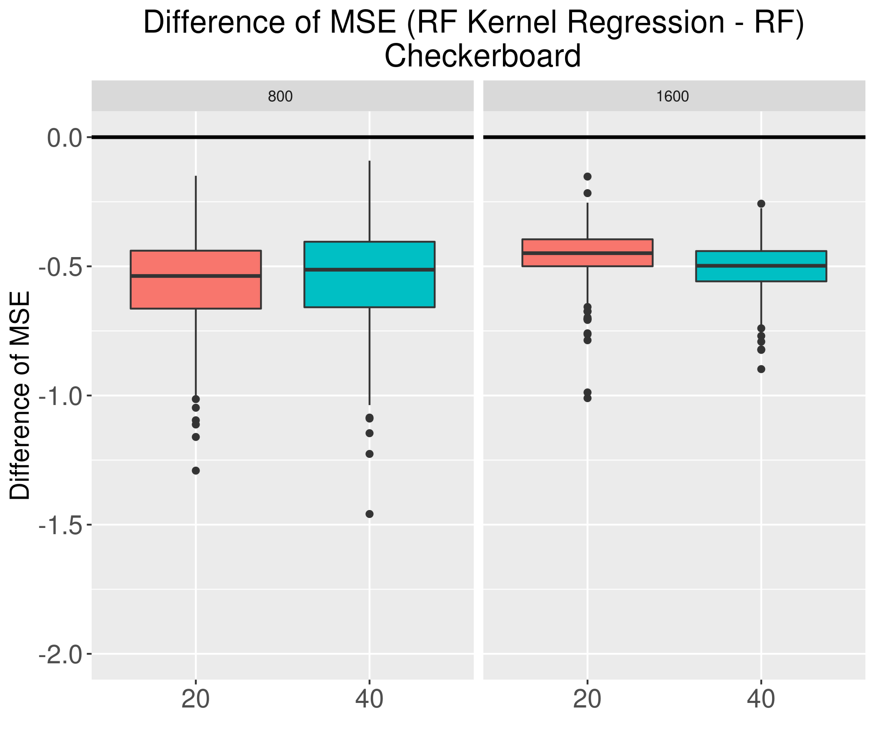}  
  \caption{Continuous Difference of MSE}
  \label{fig:sub-first}
\end{subfigure}
\begin{subfigure}{.33\textwidth}
  \centering
  \includegraphics[height=0.2\textheight]{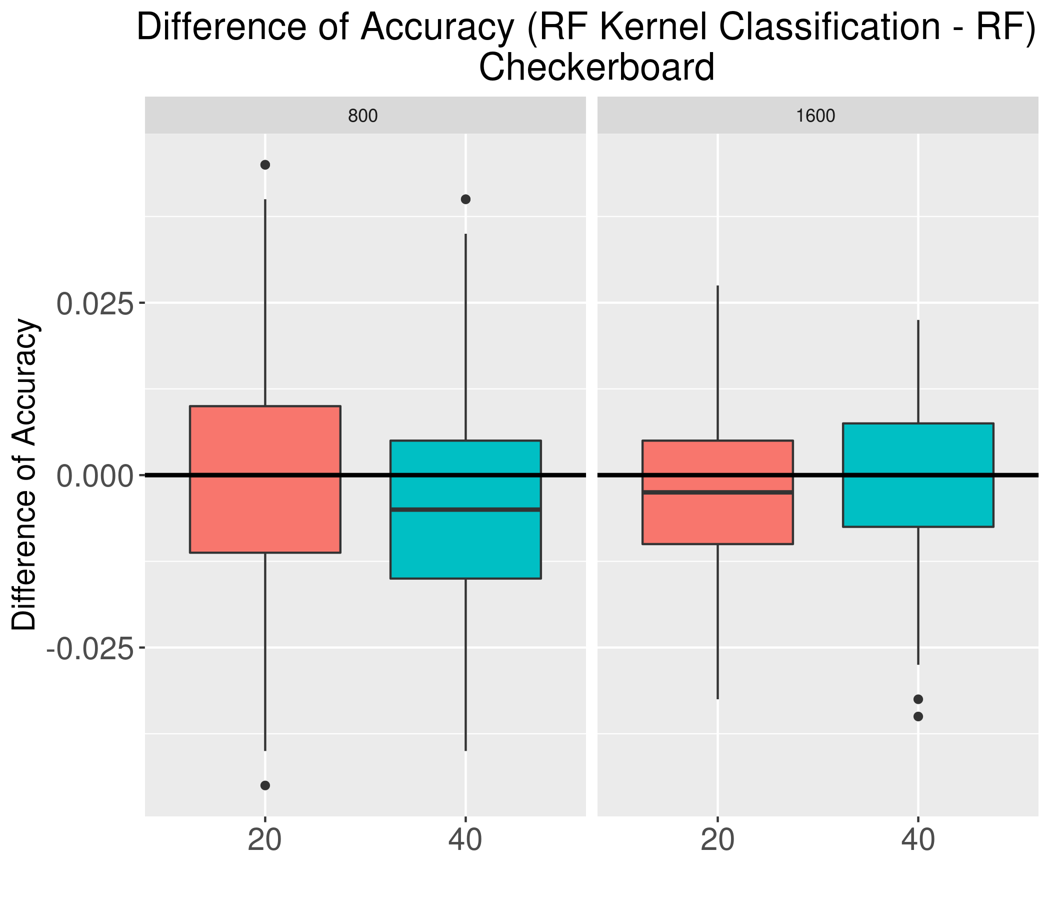} 
  \caption{Binary Difference of Accuracy}
  \label{fig:sub-second}
\end{subfigure}
\begin{subfigure}{.33\textwidth}
  \centering
  \includegraphics[height=0.2\textheight]{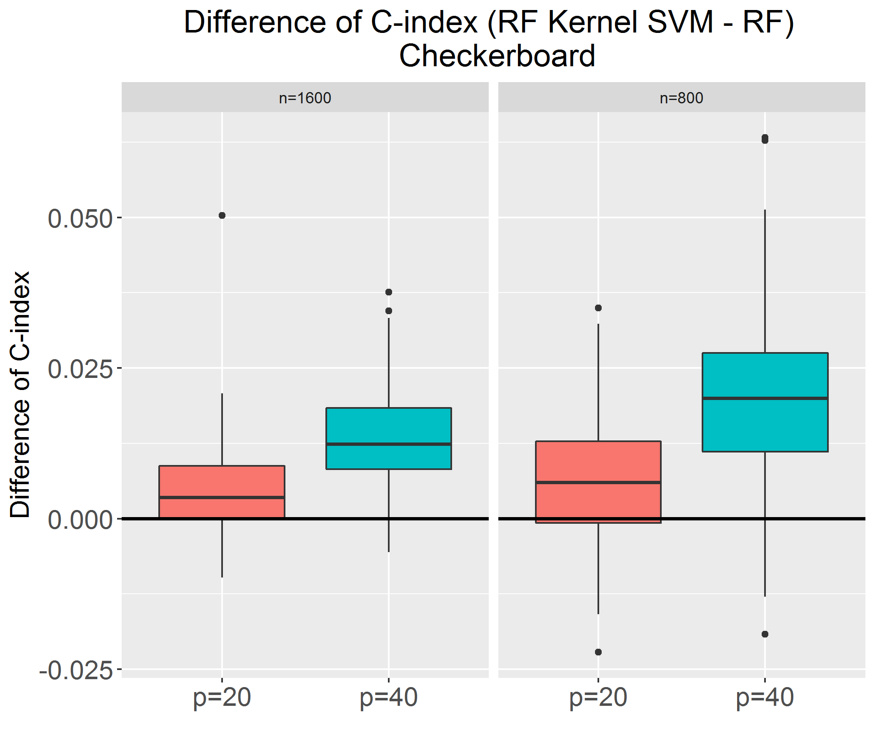} 
  \caption{Survival Difference of C-index}
  \label{fig:sub-second}
\end{subfigure}
\caption{Comparison of MSE, classification accuracy and  C-index for continuous, binary and survival targets, respectively, using RF, RF kernel and Laplace kernel for data simulated from Checkerboard setting}
\label{fig:suppCheckerboard}
\end{figure}

\begin{figure}[ht]
\begin{subfigure}{.33\textwidth}
  \centering
  \includegraphics[height=0.2\textheight]{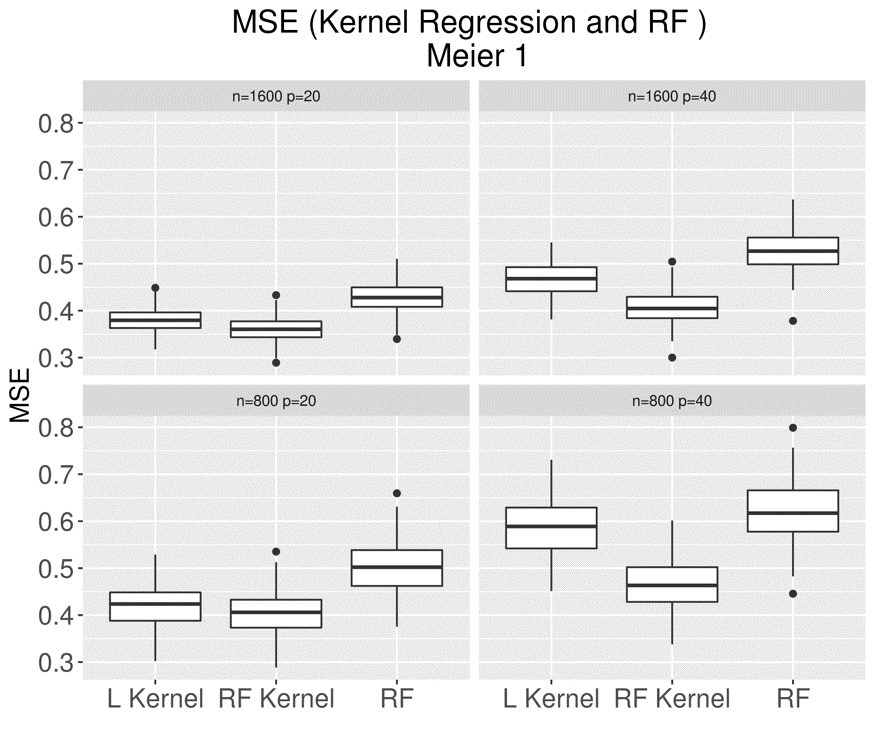}  
  \caption{Continuous MSE}
  \label{fig:sub-first}
\end{subfigure}
\begin{subfigure}{.33\textwidth}
  \centering
  \includegraphics[height=0.2\textheight]{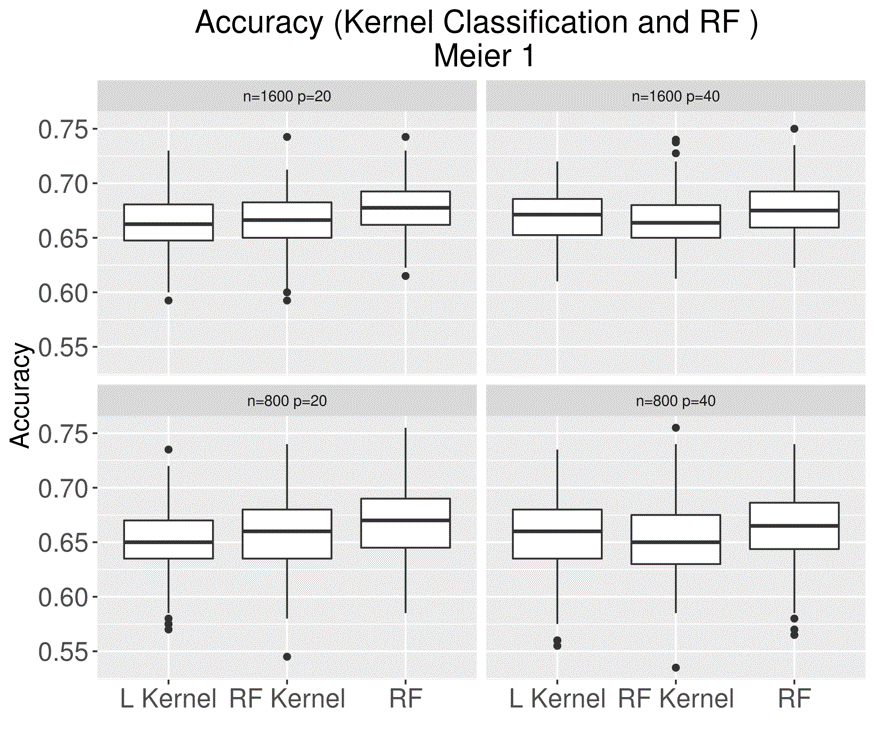}  
  \caption{Binary Accuracy}
  \label{fig:sub-first}
\end{subfigure}
\begin{subfigure}{.33\textwidth}
  \centering
  \includegraphics[height=0.2\textheight, width=0.9\textwidth]{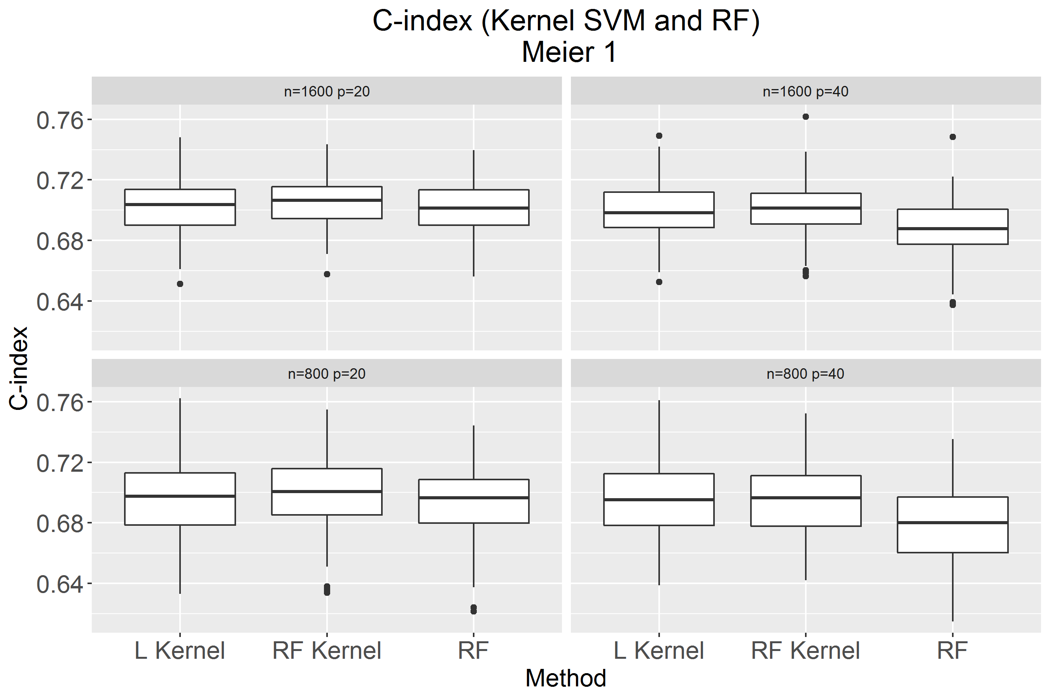}  
  \caption{Survival C-index}
  \label{fig:sub-first}
\end{subfigure}\\
\begin{subfigure}{.33\textwidth}
  \centering
  \includegraphics[height=0.2\textheight]{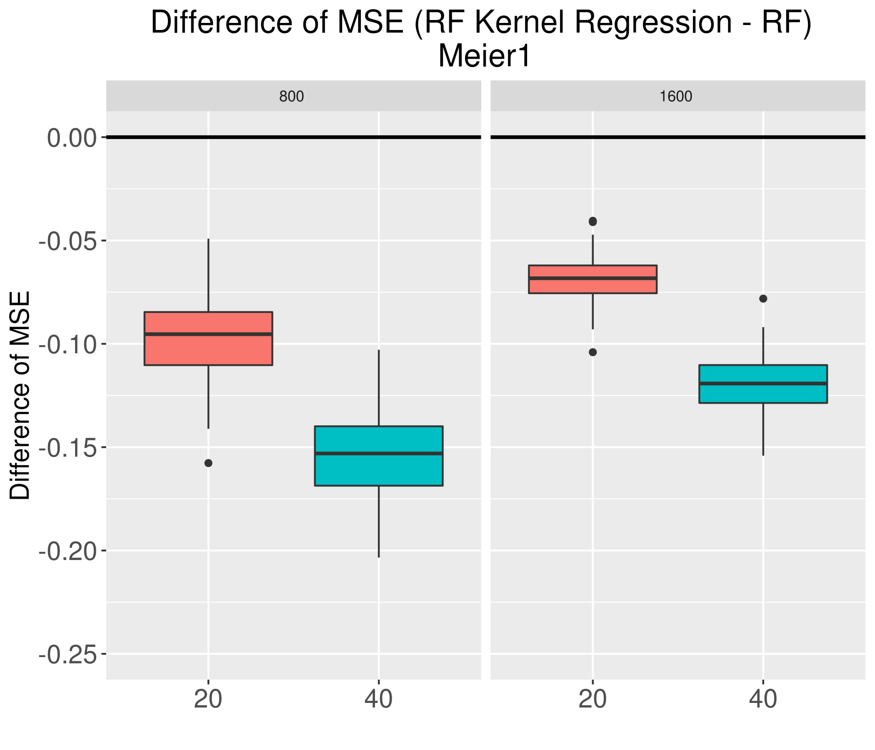}  
  \caption{Continuous Difference MSE}
  \label{fig:sub-first}
\end{subfigure}
\begin{subfigure}{.33\textwidth}
  \centering
  \includegraphics[height=0.2\textheight]{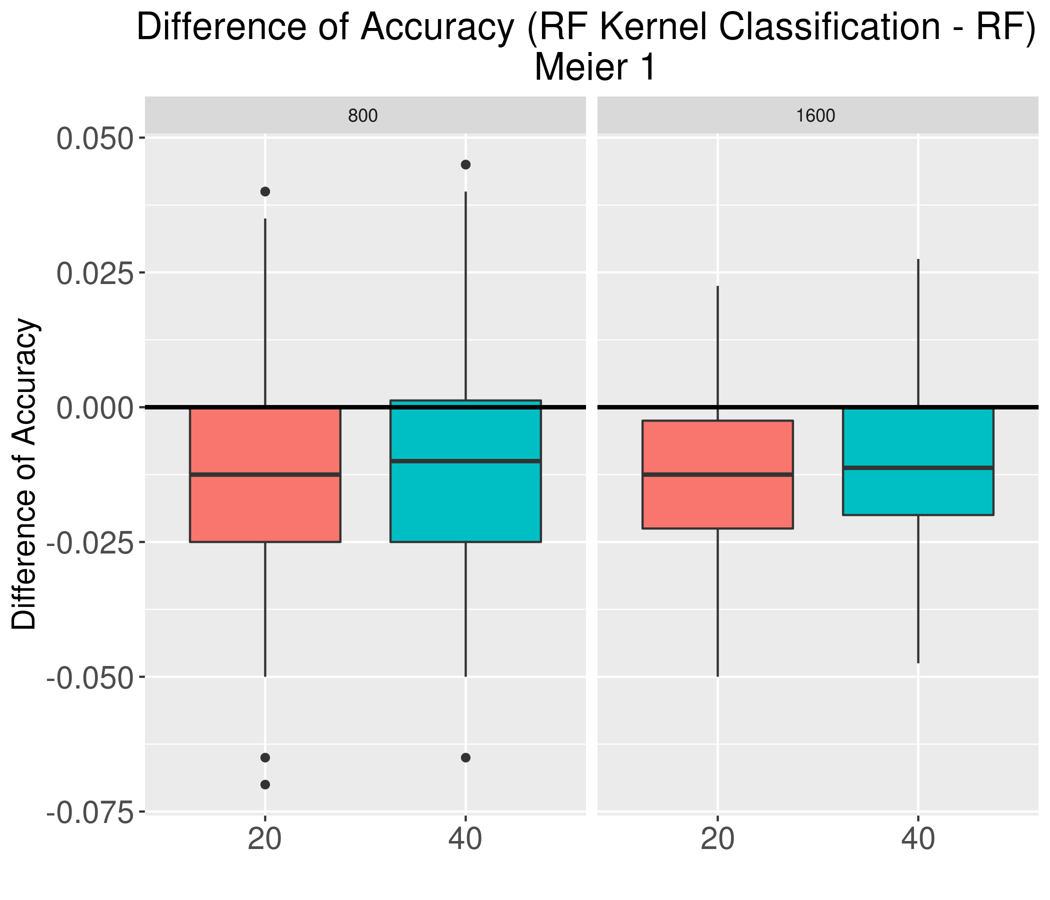} 
  \caption{Binary Difference Accuracy}
  \label{fig:sub-second}
\end{subfigure}
\begin{subfigure}{.33\textwidth}
  \centering
  \includegraphics[height=0.2\textheight]{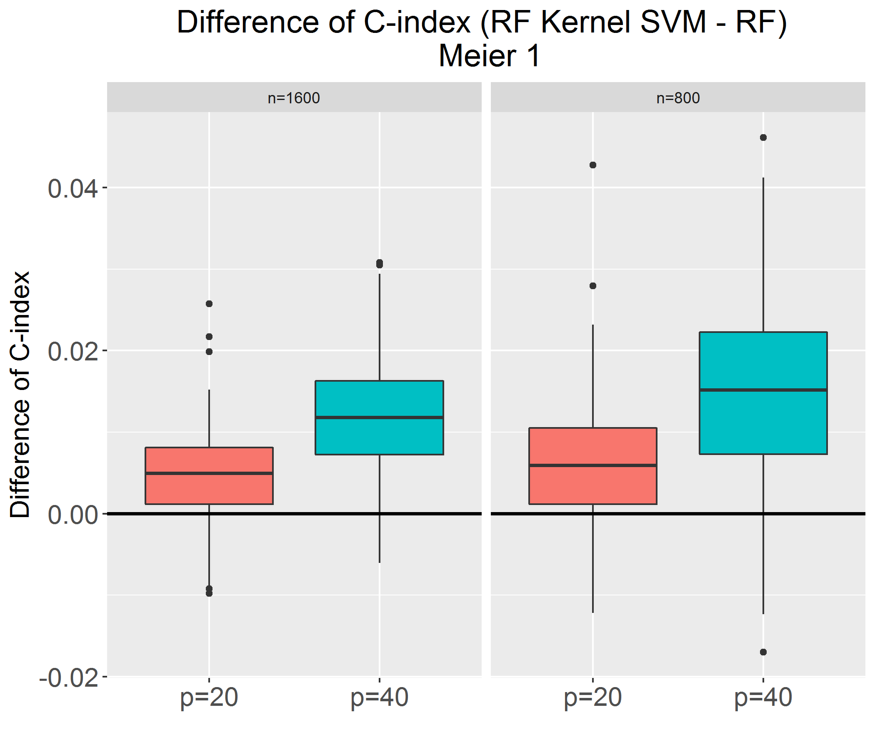} 
  \caption{Survival Difference C-index}
  \label{fig:sub-second}
\end{subfigure}
\caption{Comparison MSE, classification accuracy and  C-index for continuous, binary and survival targets, respectively, using RF, RF kernel and Laplace kernel for data simulated from Meier 1 setting}
\label{fig:suppMeier1}
\end{figure}

\begin{figure}[ht]
\begin{subfigure}{.33\textwidth}
  \centering
  \includegraphics[height=0.2\textheight]{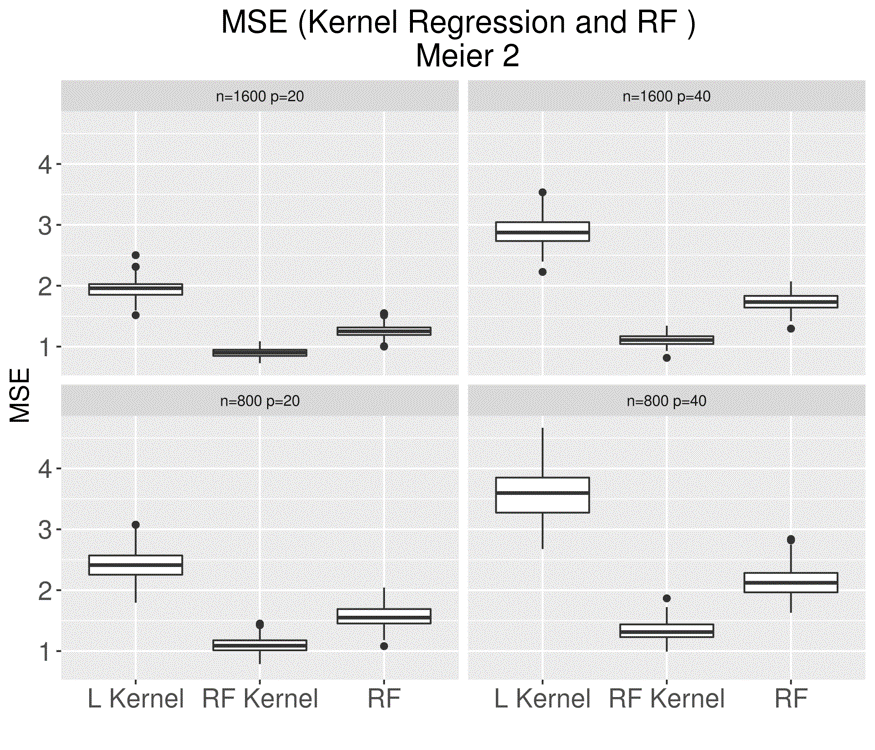}  
  \caption{Continuous MSE}
  \label{fig:sub-first}
\end{subfigure}
\begin{subfigure}{.33\textwidth}
  \centering
  \includegraphics[height=0.2\textheight]{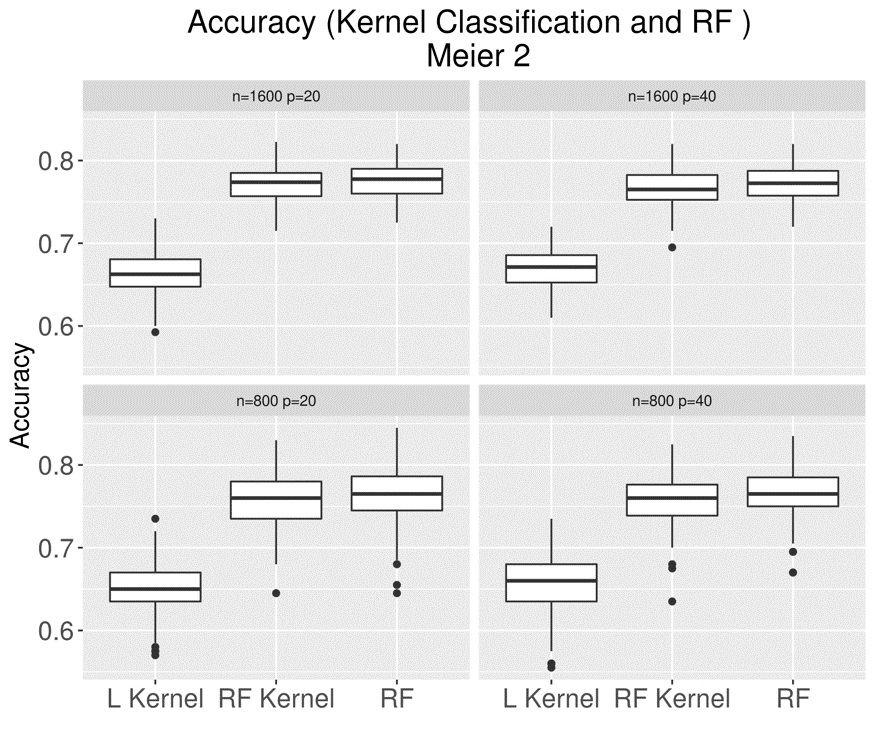}  
  \caption{Binary Accuracy}
  \label{fig:sub-first}
\end{subfigure}
\begin{subfigure}{.33\textwidth}
  \centering
  \includegraphics[height=0.2\textheight, width=0.9\textwidth]{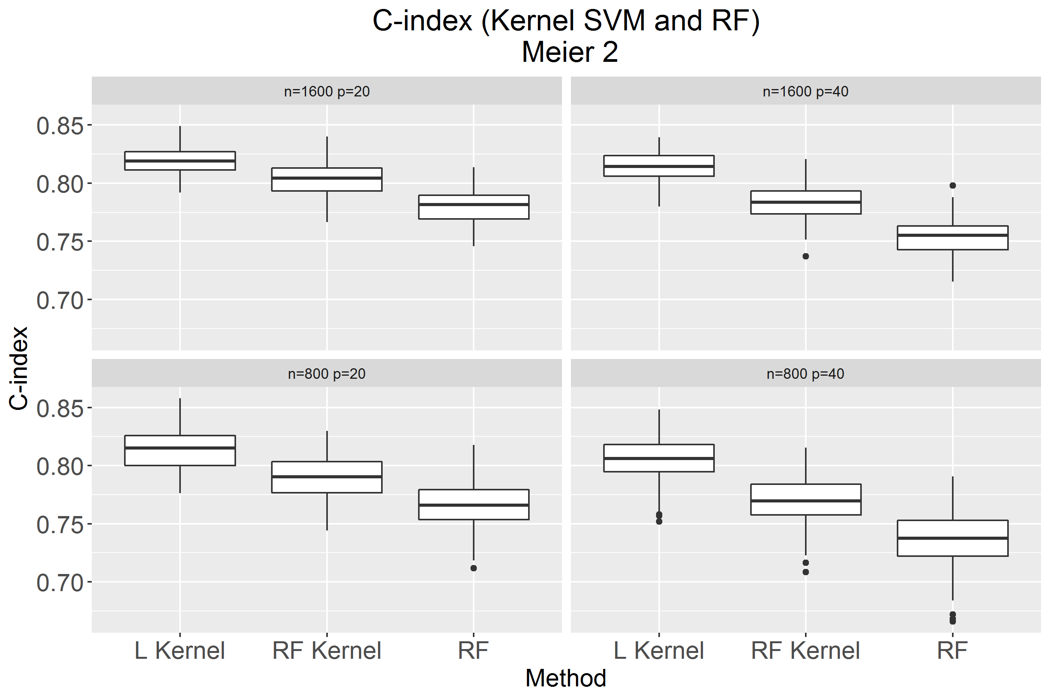}  
  \caption{Survival C-index}
  \label{fig:sub-first}
\end{subfigure}\\
\begin{subfigure}{.33\textwidth}
  \centering
  \includegraphics[height=0.2\textheight]{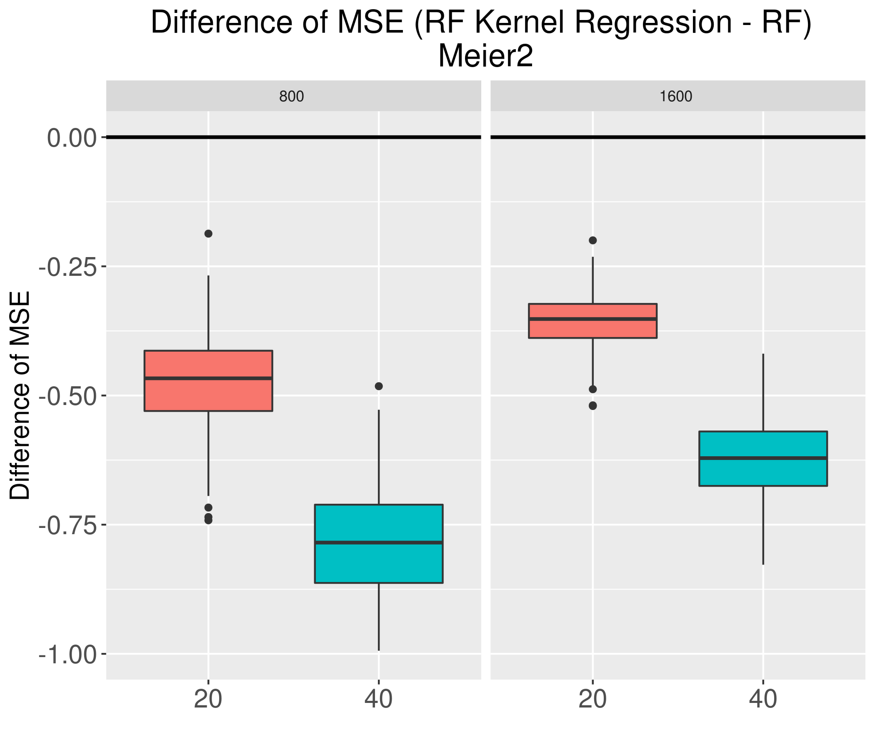}  
  \caption{Continuous Difference MSE}
  \label{fig:sub-first}
\end{subfigure}
\begin{subfigure}{.33\textwidth}
  \centering
  \includegraphics[height=0.2\textheight]{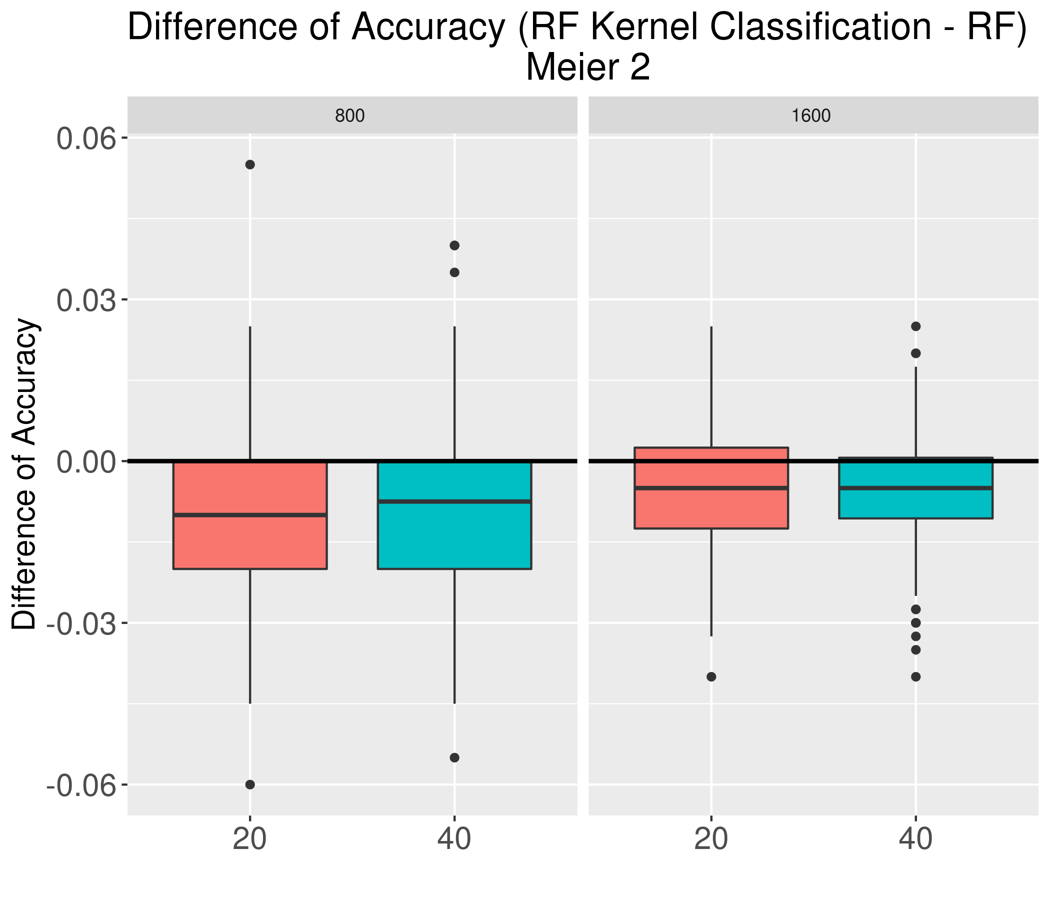} 
  \caption{Binary Difference Accuracy}
  \label{fig:sub-second}
\end{subfigure}
\begin{subfigure}{.33\textwidth}
  \centering
  \includegraphics[height=0.2\textheight]{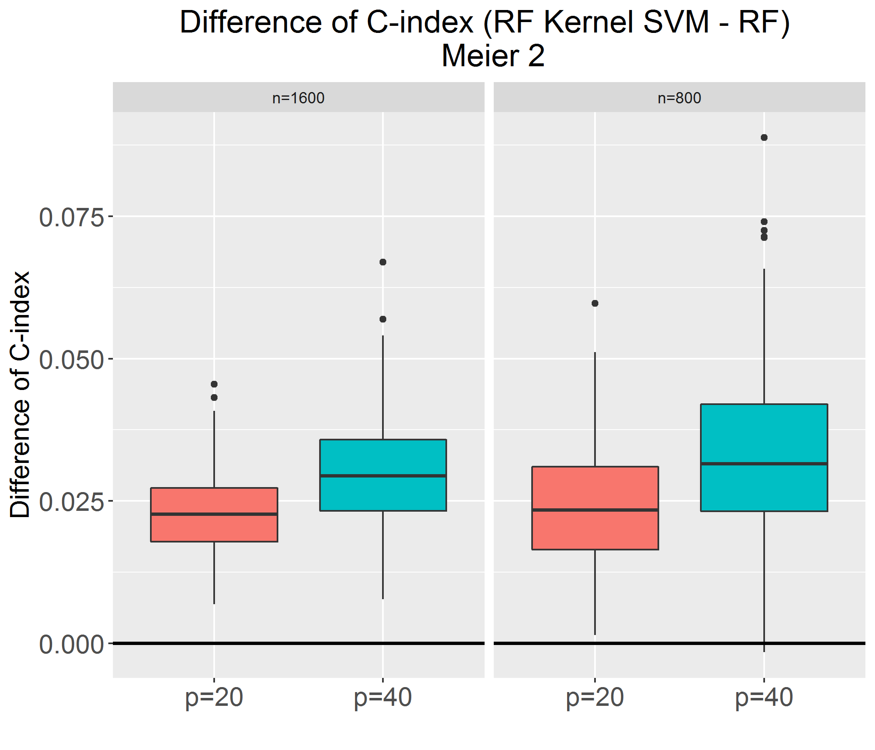} 
  \caption{Survival Difference C-index}
  \label{fig:sub-second}
\end{subfigure}
\caption{Comparison of MSE, classification accuracy and  C-index for continuous, binary and survival targets, respectively, using RF, RF kernel and Laplace kernel for data simulated from Meier 2 setting}
\label{fig:suppMeier2}
\end{figure}

\begin{figure}[ht]
\begin{subfigure}{.33\textwidth}
  \centering
  \includegraphics[height=0.2\textheight]{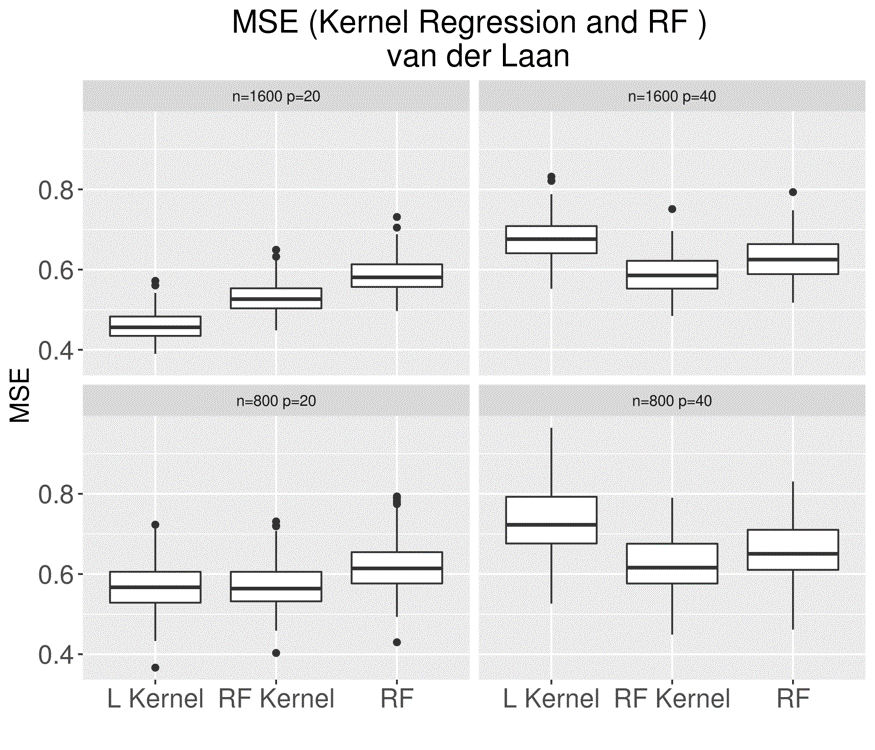}  
  \caption{Continuous MSE}
  \label{fig:sub-first}
\end{subfigure}
\begin{subfigure}{.33\textwidth}
  \centering
  \includegraphics[height=0.2\textheight]{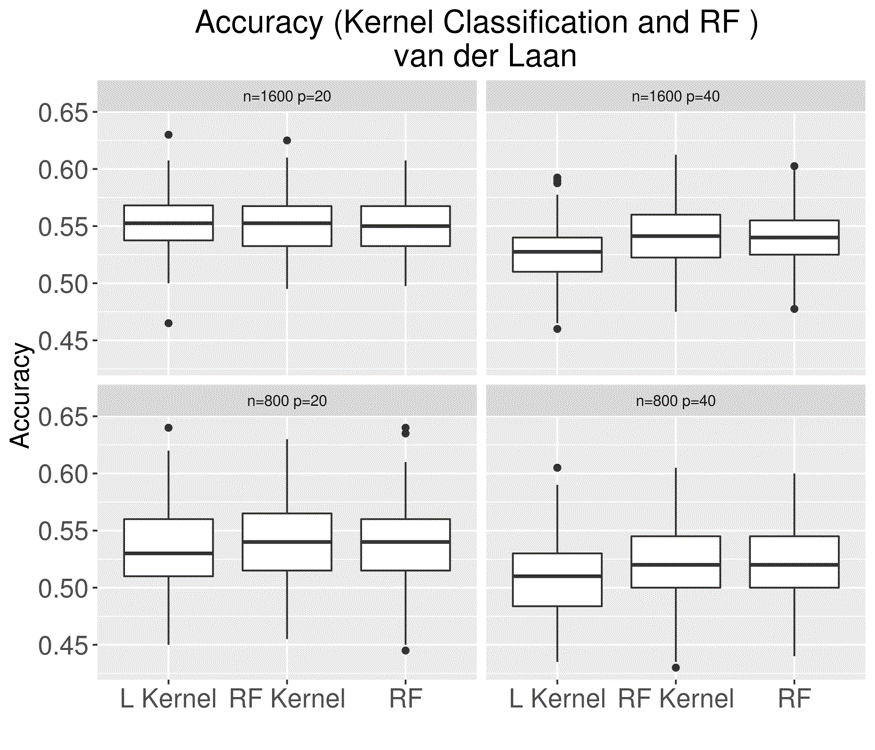}  
  \caption{Binary Accuracy}
  \label{fig:sub-first}
\end{subfigure}
\begin{subfigure}{.33\textwidth}
  \centering
  \includegraphics[height=0.2\textheight, width=0.9\textwidth]{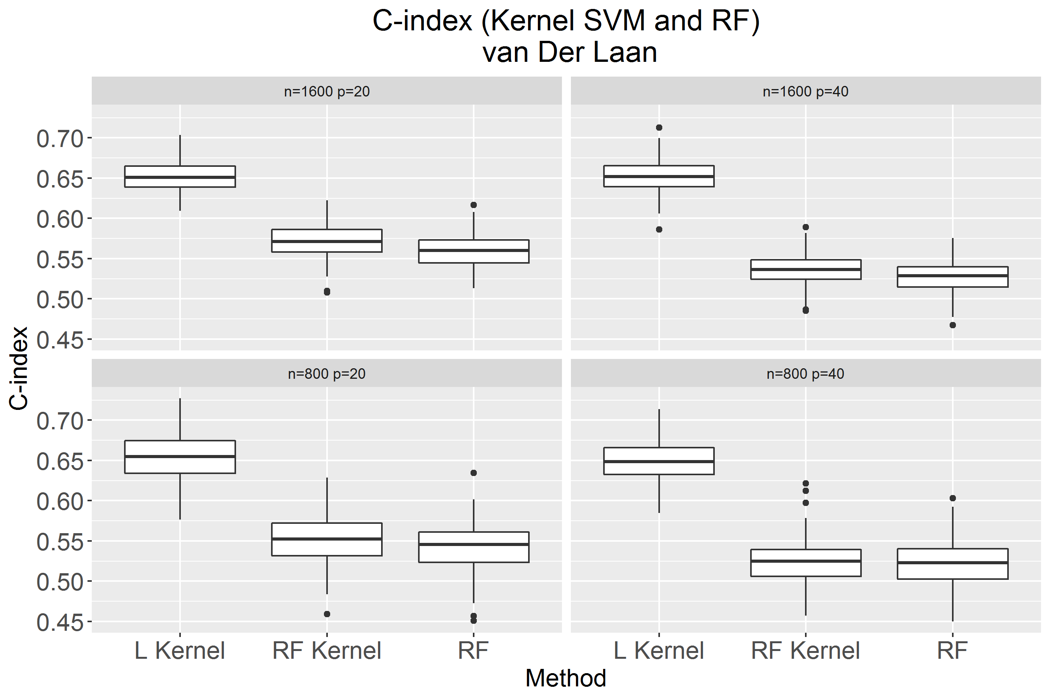}  
  \caption{Survival C-index}
  \label{fig:sub-first}
\end{subfigure}\\
\begin{subfigure}{.33\textwidth}
  \centering
  \includegraphics[height=0.2\textheight]{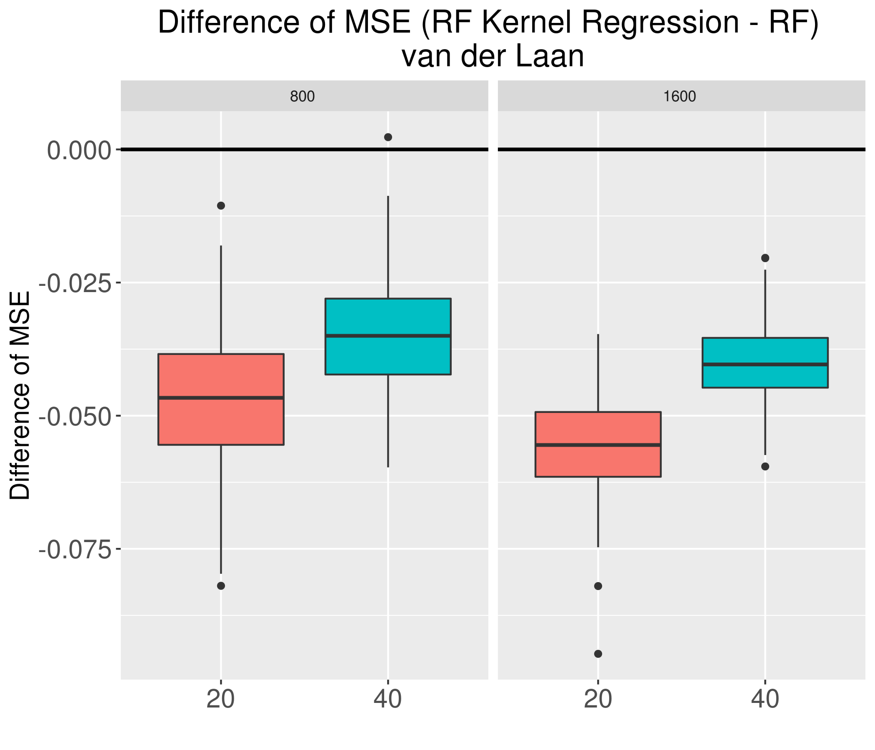}  
  \caption{Continuous Difference MSE}
  \label{fig:sub-first}
\end{subfigure}
\begin{subfigure}{.33\textwidth}
  \centering
  \includegraphics[height=0.2\textheight]{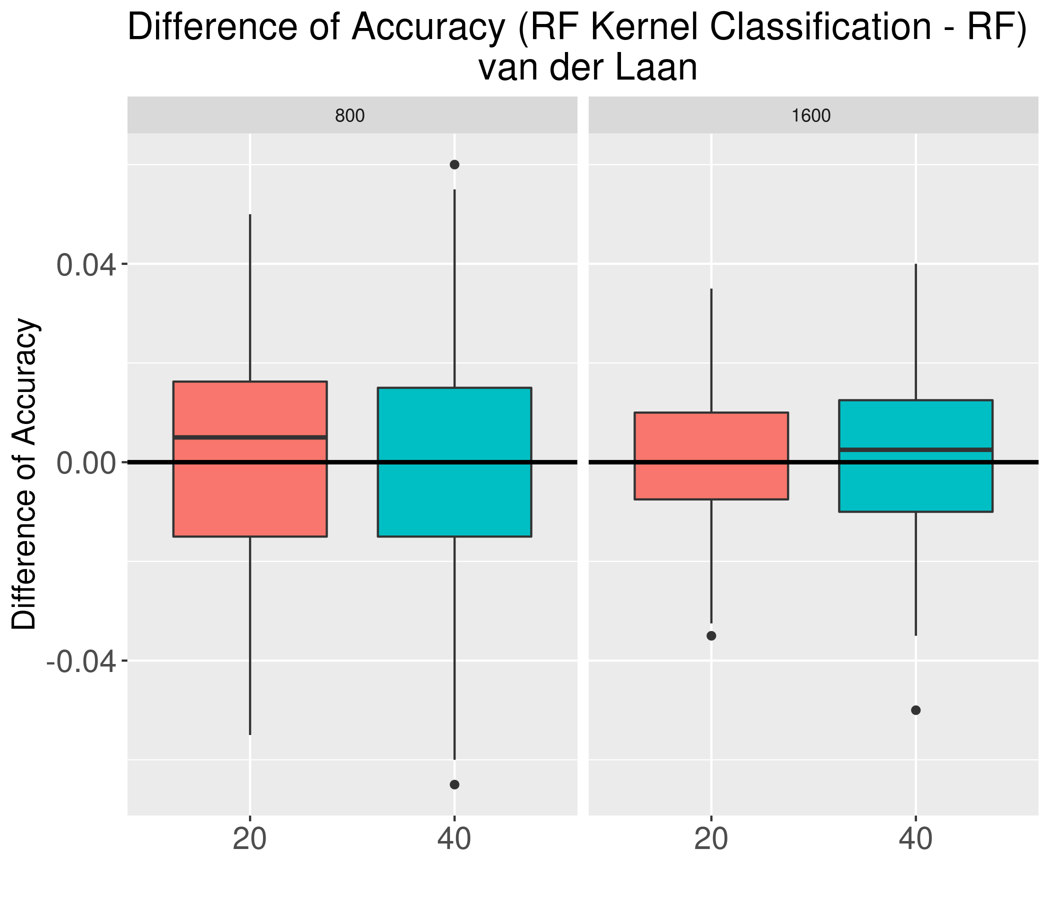} 
  \caption{Binary Difference Accuracy}
  \label{fig:sub-second}
\end{subfigure}
\begin{subfigure}{.33\textwidth}
  \centering
  \includegraphics[height=0.2\textheight]{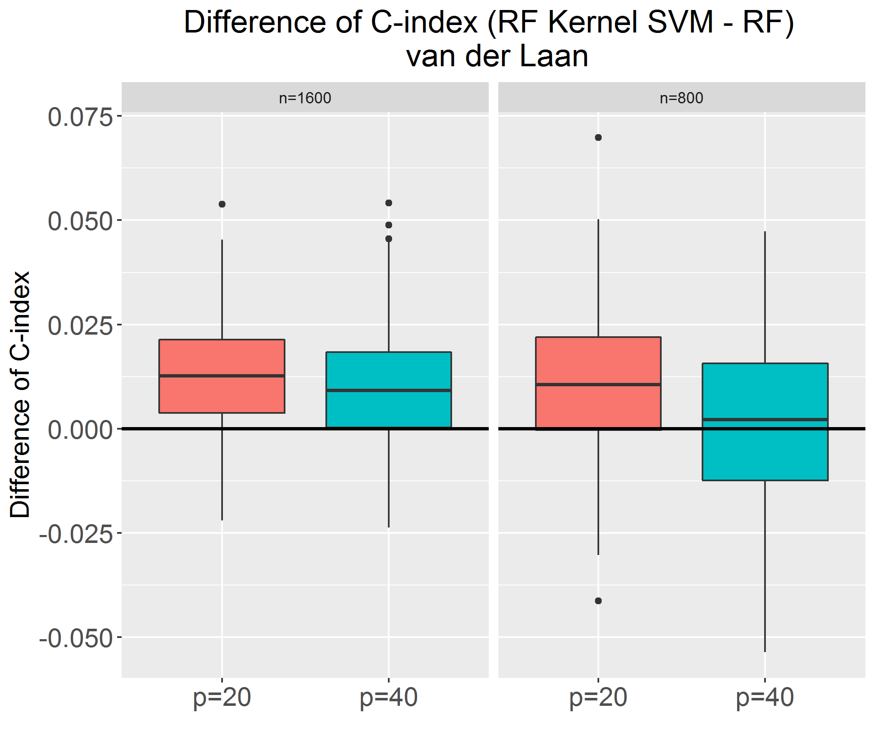} 
  \caption{Survival Difference C-index}
  \label{fig:sub-second}
\end{subfigure}
\caption{Comparison of MSE, classification accuracy and  C-index for continuous, binary and survival targets, respectively, using RF, RF kernel and Laplace kernel for data simulated from van Der Laan setting}
\label{fig:suppVanDerLaan}
\end{figure}

 The performance of the RF method, RF kernel and Laplace kernel method on test data for the Friedman generative model are shown in Figure \ref{fig:Friedman} (upper panel). To demonstrate the superiority of RF vs RF kernel, we showed also the box plots of difference in performance measures between RF kernel and RF methods in Figure \ref{fig:Friedman} (lower panel). A reference horizontal line with y-axis value equal to zero was drawn in each plot. The further away the box plot to the reference line (downward for MSE and upward for accuracy and C-index), the better the results of RF kernel compared to RF. The Figures showing the boxplots of the performance metrics differences for Checkerboard, Meier 1, Meier 2, and van der Laan are provided in Supporting Information (Figures \ref{fig:suppCheckerboard},\ref{fig:suppMeier1},\ref{fig:suppMeier2} and \ref{fig:suppVanDerLaan}). For completeness, the overall summary of the performance results across all setups for continuous, binary and survival targets are provided in Tables \ref{tab:tableContinuous}, \ref{tab:tableBinary} and \ref{tab:tableSurvival}, respectively.

From the simulation results, the RF and RF kernel outperformed the Laplace kernel for most cases and scenarios. Exception was the van der Laan case (Fig.\ref{fig:suppVanDerLaan}), where for survival the Laplace kernel performed better than RF and RF kernel (Fig.\ref{fig:suppVanDerLaan}(c)). For the regression, the Laplace kernel was competitive for the lower dimensional case ($p=20$, Fig.\ref{fig:suppVanDerLaan}(a)). In all other scenarios, the Laplace kernel performance was either incrementally worse or comparable to that of RF and RF kernel.

With respect to the RF kernel vs. RF comparison, the RF kernel generally outperformed RF for regression and survival. Furthermore, for regression and survival, it tended to be the case that with the same sample size (fixed $n$), the smaller the signal-to-noise ratio (the larger the value of $p$), the larger the improvement of adopting RF kernel approach after using RF. In addition, with a fixed number of covariates, the results from RF kernel was more accurate compared to the RF as the sample size decreased.
For classification, the performance was impacted by the target dichotomization and generally it was found comparable. Specifically, for the Friedman data the RF kernel was performing slightly better than RF (Fig.\ref{fig:Friedman}(b)), whereas for Meier 1 (Fig.\ref{fig:suppMeier1}(b)) and Meier 2 (Fig.\ref{fig:suppMeier2}(b)), the RF kernel was marginally worse. For the checkerboard (Fig.\ref{fig:suppCheckerboard}(b)) and van der Laan data (Fig.\ref{fig:suppVanDerLaan}(b)), the RF and RF kernel performances were about the same.

The results from the doubled minimum terminal node size are given in Supplementary information in Tables
\ref{tab:tableContinuousNode2},\ref{tab:tableBinaryNode2},\ref{tab:tableSurvivalNode2} for regression, classification and survival, respectively. These results were in line with those from the primary analysis.





\section{Real Data}\label{sec4}
\subsection{Regression and Classification (Continuous and binary outcome)}
\begin{figure}[h]
\centerline{\includegraphics[width=0.8\textwidth,height=0.5\textheight]{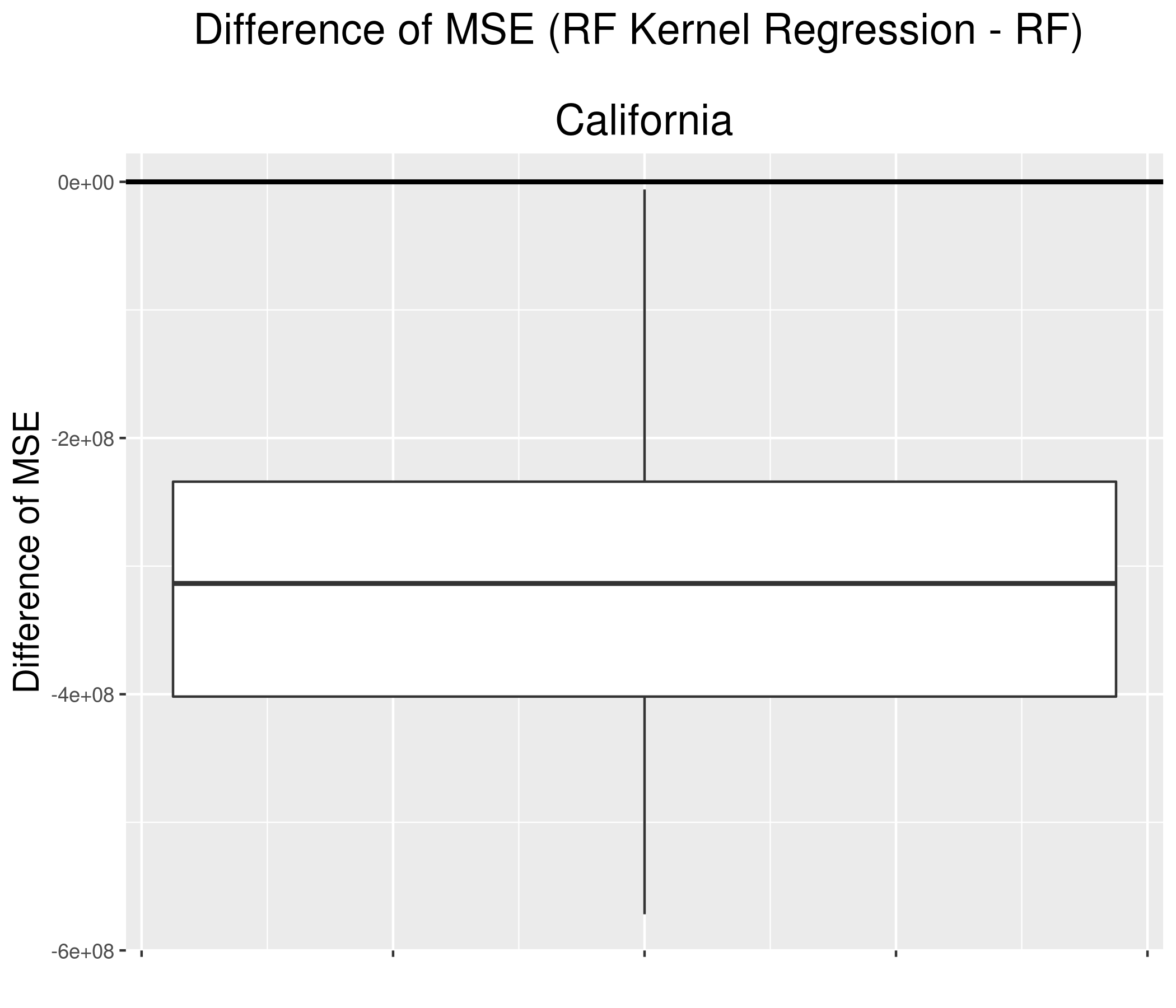}}
\caption{Comparison of MSE for the California housing data.\label{fig:housingC}}
\end{figure}

\begin{figure}[h]
\centerline{\includegraphics[width=0.8\textwidth,height=0.5\textheight]{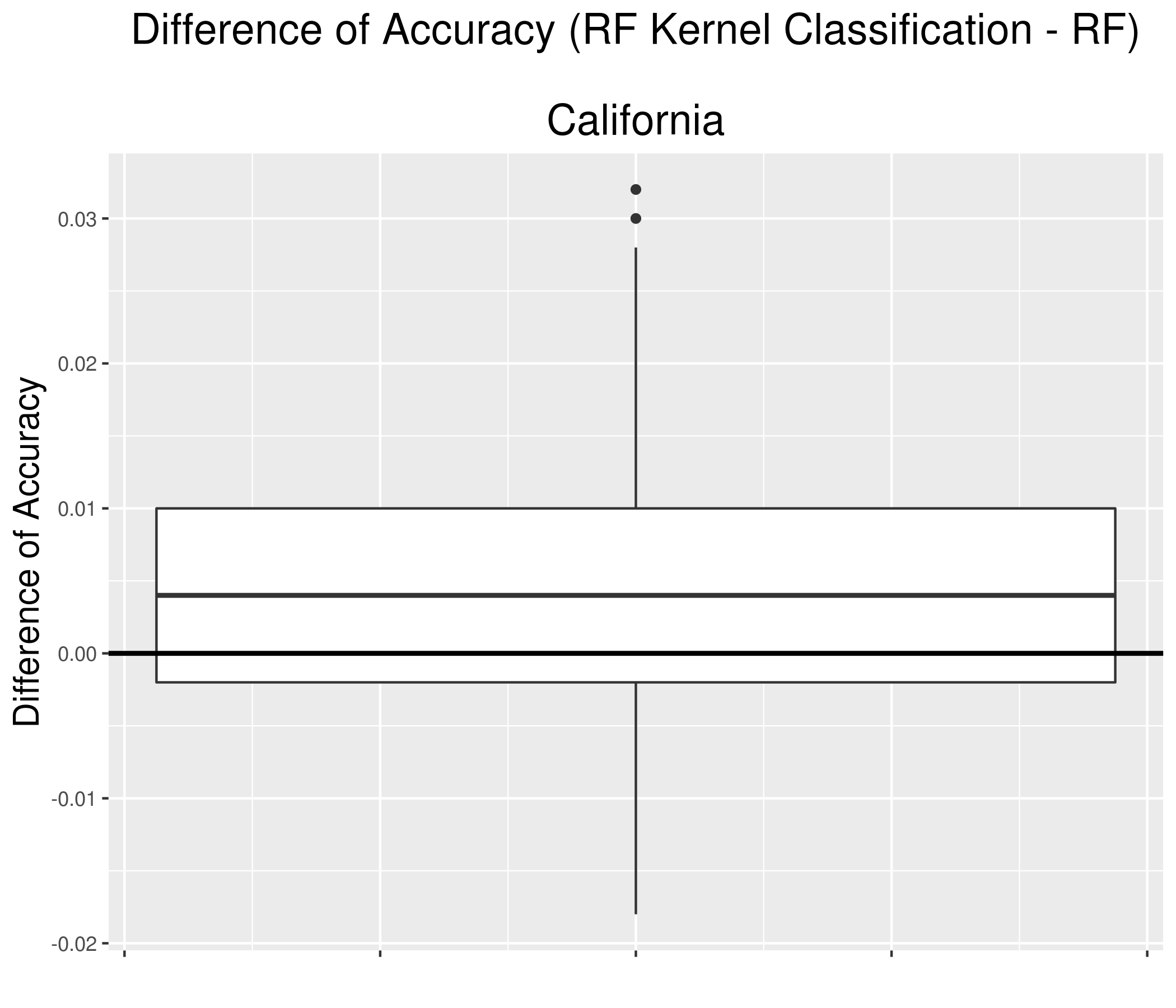}}
\caption{Comparison of classification accuracy for the California housing data.\label{fig:housingB}}
\end{figure}

To illustrate the performance of the RF kernel, we used a Kaggle California house price data set , with $n = 20640, p = 9$, to predict the median house value \cite{pace1997}. The data set was obtained from the Kaggle repository Learning Repository \url{https://www.kaggle.com/camnugent/california-housing-prices}. For the regression task we predicted the median house value as a continuous target. For classification we first dichotomized the median house value around its median that rendered low and high house price target as a two class classification problem.
For regression, we randomly selected 2000 samples and split them into training and test set, with 1500 and 500 samples, respectively. We repeated the analysis 200 times to evaluate the performance of RF and RF kernel algorithms. Similarly for classification we randomly chose 2000 samples, 1000 for each class to evaluate the performance of RF and RF kernel in a binary classification setting.
Results from this real life problem mimic those obtained in simulation. In regression as well as in classification setting, the RF kernel is competitive to that of RF as shown in Figures \ref{fig:housingC} and \ref{fig:housingB}. For regression, the mean (standard deviation) of the RF and RF kernel MSE across the 200 repeats were 
$3.64\times10^9$ ($3.96\times10^8$) and $3.32\times10^9$ ($4.0\times10^8$), respectively. The mean (standard deviation) of the MSE for the Laplace kernel was $4.2\times10^{10}$ ($1.9\times10^9$) i.e. higher than those obtained by the RF and RF kernel.
Similarly, for classification, the mean (standard deviation) of the RF and RF kernel accuracy across the 200 repeats were 
$0.85$ ($0.02$) and $0.86$ ($0.02$), respectively. The mean (standard deviation) of the accuracy for the Laplace kernel was $0.61$ ($0.02$) and it was lower than those obtained by the RF and RF kernel for this data.

\subsection{Survival: Time-to-event outcome}
The data was about the survival of breat cancer patients from German Breast Cancer Study Group 2. In this data set, $n=686, p=10$. The details of this data set can be found at:
\url{https://www.rdocumentation.org/packages/TH.data/versions/1.0-10/topics/GBSG2}.

\begin{figure}[h]
\centerline{\includegraphics[width=0.8\textwidth,height=0.5\textheight]{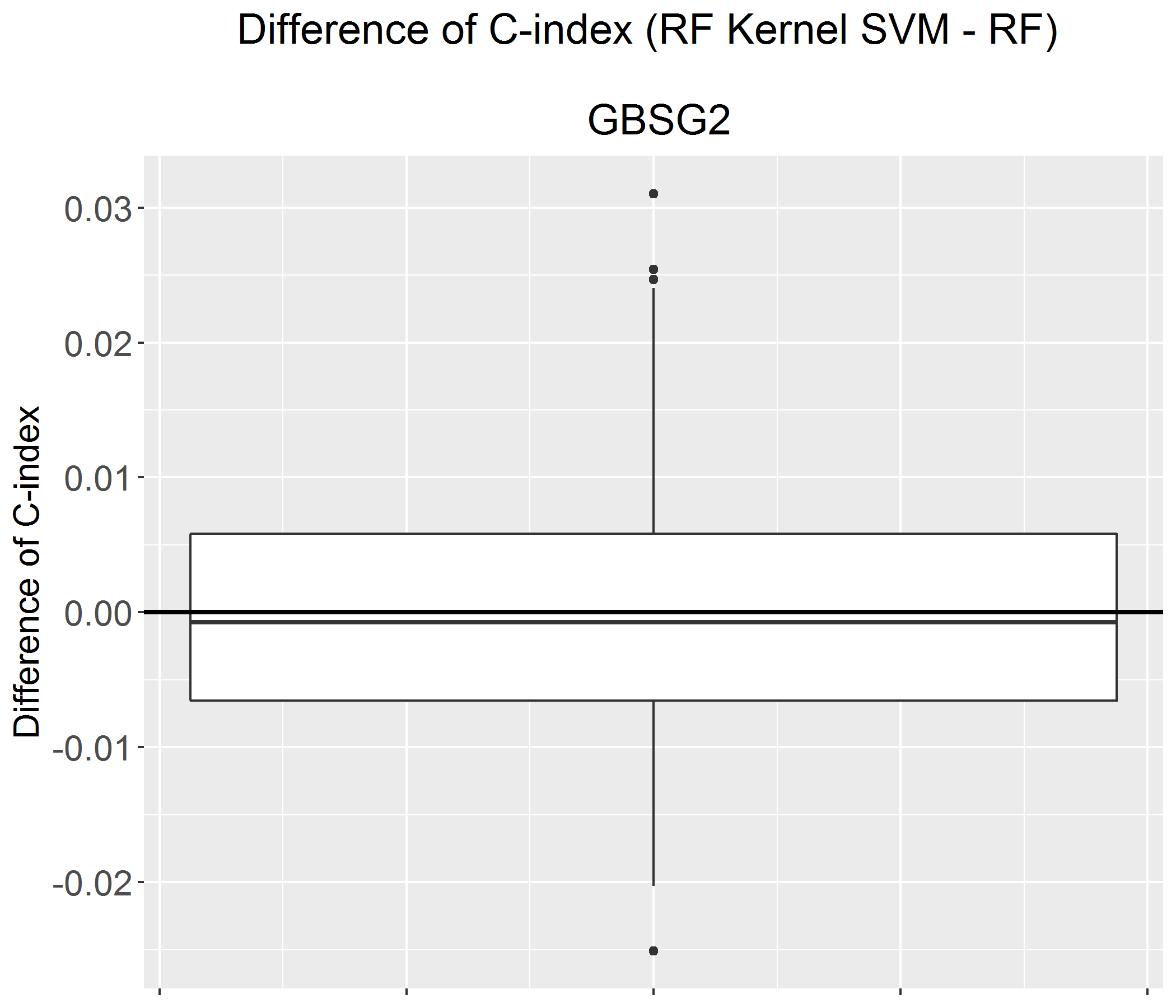}}
\caption{Comparison of C-index for real survival data.\label{fig:GBSG2}}
\end{figure}

We randomly picked 500 samples as training data and the rest was used as test data. A RF model was fit using the training data and then the C-index on the test data was calculated based on the original RF model and SVM method using the RF kernel, respectively. The process was repeated 200 times and the differences of C-index between RF and RF kernel SVM method was shown in Figure \ref{fig:GBSG2}. The results from RF and RF kernel were comparable, where the mean and standard deviation of C-index from the RF and RF kernel were equal attaining values of 0.68 (0.03). The C-index from the Laplace kernel tended to be lower with mean equal to 0.57. The corresponding standard deviation increased to 0.05.

\section{Discussion and Conclusions}\label{sec5}
It has been noticed in \cite{marcus2017}, that the RF kernel has been overlooked and underutilized in the statistical machine learning. RF kernel matrix is akin to the variable importance \cite{ishwaran2019}, as is obtained across the different prediction targets in the same form, i.e. as an $n\times n$ matrix whose entries range between 0 and 1 and represent the estimates of probability of two points being assigned to the same terminal node.
In our contribution we systematically evaluated the RF kernel prediction models in a comprehensive simulation study that included regression, classification and survival targets. Elucidation of the properties of the survival target attracted interest recently and it has been driven by real life applications \cite{ishwaran2019}.  Although the RF kernel is furnished by the RF, the prediction model is built in a different way than that of RF. The difference in these two approaches, specifically the way how they use the partitions of the data emanating from the recursive tree partitioning has been also noted in Ref. \cite{balog2016}. In \cite{balog2016} the Mondrian forest and Mondrian kernel in Bayesian framework are contrasted. It has been pointed out that the predictor of the Mondrian forest is obtained by averaging across the single trees, whereas the predictor for the Mondrian kernel is obtained jointly from the kernel by a linear learning method (model). In our case, as RF underlies both RF and RF kernel, where the prediction model for the RF kernel is a linear model that capitalizes on the RF kernel. It is expected that RF and RF kernel will be working in a similar fashion \cite{balog2016}, nevertheless there still may be differences. In our simulations we showed that for cases with larger number of noisy features the RF kernel approach may be superior to that of RF itself. This beneficial effect was found particularly consistent for the continuous and time-to-event targets, i.e. regression and survival. The simple linear model that follows the RF kernel construction (akin to the commonly used kernel methods) was found to be less prone to the noisy features in the simulation scenarios investigated. For classification we used the kernel ridge regression with target classes denoted as -1 and 1. Due to dichotomization of the continuous target \cite{fedorov2008}, the results were less pronounced for classification that those for the regression. There are other potential options for linear models that could have been used here. To this end, we also tried the regularized kernel logistic regression as implemented in the package gelnet \cite{sokolov2016} with RF and RF kernel yielding comparable performance to each other and with that of the kernel ridge regression.

The RF kernel (and RF) outperformed the Laplace kernel in our simulation study in most cases. There were scenarios still where Laplace kernel was competitive e.g. for the van der Laan data for regression and survival, demonstrating that the Laplace kernel is a valuable option to be considered in practice. There is no free lunch for statistical learning and consequently for a universally optimal kernel \cite{wolpert1996}, \cite{davies2014}, \cite{fernandezdelgado2014}. The success of a particular kernel algorithm depends on how well it adapts to the data geometry \cite{olson2018}, i.e. how well it captures the inherent kernel function of a given problem \cite{balcan2008}. The RF and accordingly the RF kernel should be competitive in situations when the data generating mechanism is conducive to the recursive partitioning, e.g. in the presence of feature interactions as frequently found in biomedical applications \cite{Boulesteix2012}. Another recent example, where the RF kernel has shown promise is a study of the image classification in hyperspectral imaging \cite{zafari2019}. Moreover, in a large bench-marking study of general purpose classification algorithms \cite{fernandezdelgado2014}, RF was found superior to other competitors. Interestingly, kernel methods that used the Gaussian kernel performed also well and were only slightly inferior to the RF. These results suggest that across broader spectrum of real life problems RF classifier adapts well to the underlying data structure \cite{olson2018} and in many cases performs better than the classifiers based on the conventional kernels such as the Gaussian kernel. It would be of interest to conduct more research into how the results from \cite{fernandezdelgado2014} extend to regression and survival and what implications they have for RF kernel and the traditionally used analytical kernels (including radial basis function, polynomial and neural network kernels \cite{friedmanHastieTibshirani2009}). 

In addition to the simulations, we have also shown that in real life applications RF kernel is competitive to RF.
However, the usefulness of RF kernel lies not only in a potential improvement of performance in certain high-dimensional setups. Availability of the RF kernel for regression, classification and survival explicitly renders the similarity/dissimilarity of the points ($\mbold{X}$-s) induced by the supervised RF kernel. This can be then straightforwardly leveraged to define prototypical (archetypal) points (observations) with insights into the geometry of a given problem. Usefulness of the prototypes has been shown for the classification in \cite{bien2011}, but the generality of the RF kernel extends it also to regression and survival. RF kernel can be also used for prototypical or landmarking classification \cite{pekalska2001},\cite{balcan2008},\cite{kar2011}.  Using this approach the similarity/dissimilarity of the points to the points in the reference/landmarking set provides for an embedding that can be used not only to achieve a competitive prediction performance but also for an improved understanding of the intrinsic dimensionality of the problem. Further research in this direction is germane to solving real world prediction problems in classification, regression and survival.

In our work we focused on the RF kernel in frequentist framework. RF kernels obtained from the the Bayesian random forests (e.g. Mondrian forests \cite{balog2016}) or Bayesian boosting by BART \cite{linero2017} can be obtained for regression, classification and survival. Therefore further understanding of the performance of the kernels from the Bayesian approaches is another interesting topic for future research. 

Besides point estimates, the RF kernel can be utilized in a Gaussian process to obtain prediction intervals to quantify the uncertainly around prediction estimates, which can be harnessed in subsequent decision making.

\section*{Acknowledgments}
We would like to thank the anonymous reviewer for his/her insightful comments and suggestions. They greatly improved the quality of the manuscript.

\subsection*{Author contributions}
All authors contributed equally to this manuscript.

\subsection*{Financial disclosure}

There is no financial disclosure to report.    

\subsection*{Conflict of interest}

The authors declare no potential conflict of interests.

\clearpage
\section*{Supporting information}
\label{sec:supp}
The following supporting information is available as part of the online article:

\noindent

\section{Figures of the Boxplots of the Difference of the Performance Metrics Across the Simulation Setups}

\noindent

\noindent

\noindent

\noindent
\clearpage

\section{Tables of Performance Results for Continuous, Binary and Survival Targets Across All
Setups}
\begin{table}
\centering
   \caption{Summary of simulation results of the mean squared error (MSE) for continuous target}

\begin{tabular}{rlrrrrrr}
  \hline
  & Setup & n & p & RF & RF kernel & L kernel & $\Delta_{RF}$  \\ 
&       &   &   & mean (sd) & mean (sd) & mean (sd) & mean (sd)  \\
  \hline

1 & Friedman & 800 & 20 &6.827 (0.668)&5.233 (0.597)&6.709 (0.793)&-1.594 (0.34)\\
2 & Friedman & 800 & 40 &8.931 (0.857)&6.558 (0.718)&10.465 (1.175)&-2.373 (0.394)\\
3 & Friedman & 1600& 20 & 5.548 (0.402)&4.303 (0.352)&5.266 (0.438)&-1.245 (0.204)\\
4 & Friedman & 1600& 40 &7.419 (0.523)&5.45 (0.411)&8.353 (0.604)&-1.969 (0.254)\\
5 & checkerboard&800&20&3.805 (0.847)&3.24 (0.771)&8.992 (3.022)&-0.565 (0.195)\\
6 & checkerboard&800&40&4.033 (0.863)&3.491 (0.751)&15.52 (3.386)&-0.543 (0.208)\\
7 & checkerboard&1600&20&3.230 (0.498)&2.77 (0.444)&6.751 (1.795)&-0.460 (0.117)\\
8 & checkerboard&1600&40&3.481 (0.564)&2.973 (0.506)&13.945 (2.58)&-0.508 (0.109)\\
9 & Meier 1&800&20&0.500 (0.054)&0.403 (0.042)&0.421 (0.043)&-0.097 (0.019)\\
10 & Meier 1&800&40&0.621 (0.065)&0.466 (0.051)&0.587 (0.061)&-0.156 (0.022)\\
11 & Meier 1&1600&20&0.429 (0.032)&0.36 (0.027)&0.381 (0.027)&-0.069 (0.01)\\
12 & Meier 1&1600&40&0.529 (0.042)&0.409 (0.033)&0.469 (0.035)&-0.120 (0.014)\\
13 & Meier 2&800&20&1.567 (0.173)&1.092 (0.127)&2.42 (0.236)&-0.475 (0.095)\\
14 & Meier 2&800&40&2.140 (0.241)&1.335 (0.153)&3.572 (0.402)&-0.806 (0.133)\\
15 & Meier 2&1600&20&1.255 (0.099)&0.9 (0.071)&1.95 (0.147)&-0.355 (0.055)\\
16 & Meier 2&1600&40&1.732 (0.135)&1.108 (0.088)&2.89 (0.225)&-0.624 (0.074)\\
17 & van der Laan&800&20&0.617 (0.061)&0.571 (0.058)&0.567 (0.055)&-0.046 (0.013)\\
18 & van der Laan&800&40&0.658 (0.069)&0.623 (0.064)&0.729 (0.08)&-0.035 (0.011)\\
19 & van der Laan&1600&20&0.586 (0.043)&0.53 (0.039)&0.459 (0.035)&-0.055 (0.009)\\
20 & van der Laan&1600&40&0.628 (0.048)&0.588 (0.045)&0.674 (0.049)&-0.04 (0.008)\\
   \hline
\end{tabular}
\label{tab:tableContinuous}
\end{table}

\begin{table}
\centering
   \caption{Summary of simulation results of the classification accuracy  for binary target}
\begin{tabular}{rlrrrrrr}
  \hline
 & Setup & n & p & RF & RF kernel & L kernel & $\Delta_{RF}$  \\ 
&       &   &   & mean (sd) & mean (sd) & mean (sd) & mean (sd)  \\
  \hline
1 & Friedman & 800 & 20&0.864 (0.026)&0.871 (0.027)&0.756 (0.031)&0.006 (0.013) \\
2 & Friedman & 800 & 40&0.852 (0.028)&0.864 (0.027)&0.68 (0.052)&0.012 (0.015) \\
3 & Friedman &1600&20&0.879 (0.016)&0.881 (0.016)&0.783 (0.02)&0.002 (0.009) \\
4 & Friedman &1600&40&0.871 (0.018)&0.877 (0.017)&0.719 (0.043)&0.007 (0.01) \\
5 & Checkerboard&800&20&0.731 (0.033)&0.729 (0.034)&0.735 (0.035)&-0.001 (0.016) \\
6 & Checkerboard&800&40&0.729 (0.033)&0.726 (0.033)&0.716 (0.033)&-0.003 (0.016) \\
7 & Checkerboard&1600&20&0.740 (0.023)&0.739 (0.023)&0.743 (0.022)&-0.001 (0.011) \\
8 & Checkerboard&1600&40&0.735 (0.021)&0.734 (0.02)&0.726 (0.021)&-0.001 (0.011) \\
9 & Meier 1&800&20&0.669 (0.033)&0.658 (0.034)&0.652 (0.031)&-0.011 (0.019) \\
10 & Meier 1&800&40&0.663 (0.033)&0.652 (0.034)&0.657 (0.034)&-0.011 (0.018) \\
11 & Meier 1&1600&20&0.677 (0.024)&0.664 (0.025)&0.663 (0.025)&-0.013 (0.013) \\
12 & Meier 1&1600&40&0.676 (0.024)&0.666 (0.024)&0.669 (0.024)&-0.01 (0.015) \\
13 & Meier 2&800&20&0.766 (0.034)&0.758 (0.033)&0.74 (0.033)&-0.008 (0.016) \\
14 & Meier 2&800&40&0.766 (0.029)&0.758 (0.029)&0.738 (0.032)&-0.009 (0.017) \\
15 & Meier 2&1600&20&0.776 (0.02)&0.771 (0.02)&0.749 (0.021)&-0.005 (0.011) \\
16 & Meier 2&1600&40&0.773 (0.021)&0.767 (0.023)&0.749 (0.022)&-0.005 (0.011) \\
17 & van der Laan&800&20&0.537 (0.037)&0.54 (0.035)&0.535 (0.036)&0.003 (0.021) \\
18 & van der Laan&800&40&0.522 (0.033)&0.522 (0.034)&0.508 (0.034)&0.001 (0.022) \\
19 & van der Laan&1600&20&0.551 (0.024)&0.551 (0.025)&0.554 (0.025)&0 (0.014) \\
20 & van der Laan&1600&40&0.539 (0.023)&0.541 (0.026)&0.526 (0.023)&0.001 (0.016) \\
   \hline
\end{tabular}
\label{tab:tableBinary}
\end{table}

\begin{table}
\centering
   \caption{Summary of simulation results of the C-index  for survival target}
\begin{tabular}{rlrrrrrr}
  \hline
 & Setup & n & p & RF & RF kernel & L kernel & $\Delta_{RF}$  \\ 
 &       &   &   & mean (sd) & mean (sd) & mean (sd) & mean (sd) \\
  \hline
1 & Friedman & 800 &  20 & 0.650 (0.025) & 0.659 (0.023) & 0.616 (0.026) & 0.009 (0.011) \\ 
  2 & Friedman & 800 &  40 & 0.630 (0.028) & 0.646 (0.026) & 0.591 (0.026) & 0.016 (0.014) \\ 
  3 & Friedman & 1600 &  20 & 0.661 (0.017) & 0.668 (0.017) & 0.630 (0.018) & 0.007 (0.006) \\ 
  4 & Friedman & 1600 &  40 & 0.644 (0.017) & 0.657 (0.017) & 0.609 (0.021) & 0.012 (0.008) \\ 
  5 & Checkerboard & 800 &  20 & 0.776 (0.020) & 0.782 (0.02) & 0.766 (0.017) & 0.006 (0.01) \\ 
  6 & Checkerboard & 800 &  40 & 0.751 (0.021) & 0.771 (0.019) & 0.751 (0.021) & 0.02 (0.014) \\ 
  7 & Checkerboard & 1600 &  20 & 0.788 (0.014) & 0.792 (0.013) & 0.767 (0.012) & 0.004 (0.007) \\ 
  8 & Checkerboard & 1600 &  40 & 0.769 (0.016) & 0.782 (0.015) & 0.754 (0.014) & 0.013 (0.008) \\ 
  9 & Meier 1 & 800 &  20 & 0.694 (0.022) & 0.7 (0.023) & 0.696 (0.025) & 0.006 (0.008) \\ 
  10 & Meier 1 & 800 &  40 & 0.68 (0.025) & 0.695 (0.023) & 0.695 (0.024) & 0.015 (0.01) \\ 
  11 & Meier 1 & 1600 &  20 & 0.701 (0.015) & 0.706 (0.015) & 0.701 (0.017) & 0.005 (0.005) \\ 
  12 & Meier 1 & 1600 &  40 & 0.689 (0.017) & 0.701 (0.016) & 0.700 (0.018) & 0.012 (0.007) \\ 
  13 & Meier 2 & 800 &  20 & 0.766 (0.019) & 0.790 (0.018) & 0.814 (0.017) & 0.024 (0.011) \\ 
  14 & Meier 2 & 800 &  40 & 0.737 (0.023) & 0.770 (0.019) & 0.806 (0.019) & 0.033 (0.015) \\ 
  15 & Meier 2 & 1600 &  20 & 0.780 (0.015) & 0.803 (0.014) & 0.82 (0.012) & 0.023 (0.007) \\ 
  16 & Meier 2 & 1600 &  40 & 0.754 (0.015) & 0.783 (0.013) & 0.814 (0.013) & 0.03 (0.009) \\ 
  17 & van Der Laan & 800 &  20 & 0.543 (0.028) & 0.553 (0.03) & 0.655 (0.027) & 0.01 (0.017) \\ 
  18 & van Der Laan & 800 &  40 & 0.521 (0.028) & 0.523 (0.028) & 0.649 (0.026) & 0.002 (0.02) \\ 
  19 & van Der Laan & 1600 &  20 & 0.559 (0.021) & 0.572 (0.02) & 0.652 (0.018) & 0.013 (0.013) \\ 
  20 & van Der Laan & 1600 &  40 & 0.527 (0.021) & 0.536 (0.02) & 0.652 (0.019) & 0.009 (0.014) \\    \hline
\end{tabular}
\label{tab:tableSurvival}
\end{table}

\clearpage
\subsection{Tables of Performance Results for Continuous, Binary and Survival Targets Across All Setups for the Doubled Minimum Node Size}

\begin{table}
\centering
   \caption{Summary of simulation results of the mean squared error (MSE) for continuous target (doubled minimum node size)}
\begin{tabular}{rlrrrrr}
  \hline
 & Setup & n & p & RF & RF kernel & $\Delta_{RF}$  \\ 
 &       &   &   & mean (sd) & mean (sd) & mean (sd) \\
 \hline
 1&Friedman&800&20&6.956 (0.684)&4.677 (0.549)&-2.279 (0.363)\\
2&Friedman&800&40&9.031 (0.867)&5.899 (0.636)&-3.133 (0.432)\\
3&Friedman&1600&20&5.644 (0.407)&3.81 (0.308)&-1.834 (0.224)\\
4&Friedman&1600&40&7.49 (0.529)&4.875 (0.371)&-2.615 (0.297)\\
5&Checkerboard&800&20&3.95 (0.898)&3.097 (0.708)&-0.852 (0.3)\\
6&Checkerboard&800&40&4.129 (0.899)&3.328 (0.686)&-0.801 (0.317)\\
7&Checkerboard&1600&20&3.314 (0.518)&2.632 (0.405)&-0.682 (0.18)\\
8&Checkerboard&1600&40&3.547 (0.584)&2.807 (0.46)&-0.740 (0.181)\\
9&Meier 1&800&20&0.507 (0.055)&0.389 (0.04)&-0.118 (0.023)\\
10&Meier 1&800&40&0.627 (0.066)&0.445 (0.049)&-0.182 (0.026)\\
11&Meier 1&1600&20&0.434 (0.033)&0.35 (0.026)&-0.083 (0.012)\\
12&Meier 1&1600&40&0.533 (0.043)&0.393 (0.032)&-0.140 (0.016)\\
13&Meier 2&800&20&1.607 (0.177)&0.999 (0.117)&-0.608 (0.108)\\
14&Meier 2&800&40&2.183 (0.25)&1.229 (0.14)&-0.954 (0.157)\\
15&Meier 2&1600&20&1.284 (0.104)&0.815 (0.063)&-0.469 (0.064)\\
16&Meier 2&1600&40&1.759 (0.136)&1.004 (0.078)&-0.754 (0.09)\\
17&van der Laan&800&20&0.62 (0.062)&0.562 (0.057)&-0.058 (0.016)\\
18&van der Laan&800&40&0.659 (0.069)&0.618 (0.064)&-0.041 (0.015)\\
19&van der Laan&1600&20&0.588 (0.043)&0.519 (0.039)&-0.07 (0.012)\\
20&van der Laan&1600&40&0.628 (0.048)&0.58 (0.044)&-0.048 (0.01)\\ 
   \hline
\end{tabular}
\label{tab:tableContinuousNode2}
\end{table}

\begin{table}
\centering
   \caption{Summary of simulation results of the classification accuracy for binary target (doubled minimum node size)}
\begin{tabular}{rlrrrrr}
  \hline
 & Setup & n & p & RF & RF kernel & $\Delta_{RF}$  \\ 
 &       &   &   & mean (sd) & mean (sd) & mean (sd) \\
\hline
1&Friedman&800&20&0.865 (0.027)&0.87 (0.026)&0.005 (0.014)\\
2&Friedman&800&40&0.852 (0.028)&0.863 (0.027)&0.011 (0.015)\\
3&Friedman&1600&20&0.879 (0.016)&0.882 (0.016)&0.003 (0.008)\\
4&Friedman&1600&40&0.871 (0.018)&0.877 (0.016)&0.006 (0.01)\\
5&Checkerboard&800&20&0.732 (0.033)&0.729 (0.033)&-0.003 (0.016)\\
6&Checkerboard&800&40&0.729 (0.033)&0.726 (0.032)&-0.003 (0.015)\\
7&Checkerboard&1600&20&0.741 (0.023)&0.739 (0.023)&-0.002 (0.011)\\
8&Checkerboard&1600&40&0.737 (0.02)&0.735 (0.02)&-0.001 (0.012)\\
9&Meier 1&800&20&0.669 (0.032)&0.658 (0.034)&-0.012 (0.019)\\
10&Meier 1&800&40&0.662 (0.034)&0.649 (0.036)&-0.013 (0.019)\\
11&Meier 1&1600&20&0.677 (0.023)&0.664 (0.023)&-0.013 (0.013)\\
12&Meier 1&1600&40&0.676 (0.023)&0.667 (0.024)&-0.01 (0.014)\\
13&Meier 2&800&20&0.767 (0.033)&0.758 (0.034)&-0.008 (0.017)\\
14&Meier 2&800&40&0.766 (0.03)&0.757 (0.03)&-0.01 (0.016)\\
15&Meier 2&1600&20&0.777 (0.02)&0.771 (0.02)&-0.005 (0.011)\\
16&Meier 2&1600&40&0.773 (0.021)&0.768 (0.022)&-0.005 (0.011)\\
17&van der Laan&800&20&0.536 (0.037)&0.539 (0.035)&0.003 (0.021)\\
18&van der Laan&800&40&0.52 (0.034)&0.523 (0.034)&0.002 (0.021)\\
19&van der Laan&1600&20&0.552 (0.025)&0.551 (0.024)&-0.001 (0.015)\\
20&van der Laan&1600&40&0.54 (0.024)&0.539 (0.024)&0 (0.017)\\
   \hline
\end{tabular}
\label{tab:tableBinaryNode2}
\end{table}

\begin{table}
\centering
   \caption{Summary of simulation results of the C-index for survival target (doubled minimum node size)}
\begin{tabular}{rlrrrrr}
  \hline
 & Setup & n & p & RF & RF kernel & $\Delta_{RF}$  \\ 
 &       &   &   & mean (sd) & mean (sd) & mean (sd) \\
 \hline
1 & Friedman & 800 &  20 & 0.65 (0.025) & 0.658 (0.023) & 0.008 (0.01) \\ 
  2 & Friedman & 800 &  40 & 0.632 (0.023) & 0.647 (0.023) & 0.015 (0.013) \\ 
  3 & Friedman & 1600 &  20 & 0.662 (0.017) & 0.668 (0.017) & 0.006 (0.006) \\ 
  4 & Friedman & 1600 &  40 & 0.647 (0.017) & 0.657 (0.017) & 0.01 (0.007) \\ 
  5 & Checkerboard & 800 &  20 & 0.768 (0.02) & 0.772 (0.02) & 0.004 (0.009) \\ 
  6 & Checkerboard & 800 &  40 & 0.745 (0.023) & 0.763 (0.02) & 0.018 (0.014) \\ 
  7 & Checkerboard & 1600 &  20 & 0.783 (0.015) & 0.786 (0.013) & 0.003 (0.007) \\ 
  8 & Checkerboard & 1600 &  40 & 0.765 (0.018) & 0.776 (0.016) & 0.011 (0.009) \\ 
  9 & Meier 1 & 800 &  20 & 0.695 (0.023) & 0.7 (0.024) & 0.005 (0.008) \\ 
  10 & Meier 1 & 800 &  40 & 0.682 (0.024) & 0.694 (0.023) & 0.012 (0.011) \\ 
  11 & Meier 1 & 1600 &  20 & 0.702 (0.016) & 0.706 (0.015) & 0.004 (0.005) \\ 
  12 & Meier 1 & 1600 &  40 & 0.691 (0.017) & 0.7 (0.016) & 0.009 (0.007) \\ 
  13 & Meier 2 & 800 &  20 & 0.763 (0.02) & 0.788 (0.019) & 0.025 (0.01) \\ 
  14 & Meier 2 & 800 &  40 & 0.74 (0.022) & 0.769 (0.02) & 0.028 (0.014) \\ 
  15 & Meier 2 & 1600 &  20 & 0.777 (0.015) & 0.801 (0.015) & 0.024 (0.007) \\ 
  16 & Meier 2 & 1600 &  40 & 0.754 (0.015) & 0.782 (0.013) & 0.028 (0.009) \\ 
  17 & van Der Laan & 800 &  20 & 0.54 (0.028) & 0.548 (0.03) & 0.009 (0.017) \\ 
  18 & van Der Laan & 800 &  40 & 0.52 (0.029) & 0.523 (0.029) & 0.003 (0.016) \\ 
  19 & van Der Laan & 1600 &  20 & 0.559 (0.02) & 0.572 (0.021) & 0.013 (0.012) \\ 
  20 & van Der Laan & 1600 &  40 & 0.527 (0.021) & 0.535 (0.021) & 0.008 (0.013) \\ 
   \hline
\end{tabular}
\label{tab:tableSurvivalNode2}
\end{table}
\clearpage
\appendix

\section{Primal and Dual Survival SVM Regression Problems\label{app1}}

The primal SSVM problem can be formulated as follows:
\[
\min_{w,b,\xi,\xi^{\ast}}\frac{1}{2}w^Tw+C(\sum_{i=1}^{n}
\xi_i+\sum_{i=1}^{n}\xi_i^\ast)
\]
\[\textrm{subject to}\left \{
\begin{matrix}
w^T\phi(X_i)+b &\leq& Y_i-\xi_i  & \forall i \in {1,\dots,n}\\
-\delta_i(w^T\phi(X_i)+b) &\leq& -\delta_i Y_i -\xi_i^\ast  & \forall i \in {1,\dots,n}\\
0 &\leq& \xi_i & \forall i \in {1,\dots,n}\\
0 &\leq& \xi_i^\ast & \forall i \in {1,\dots,n}
\end{matrix}\right.
\]
where
\[
\delta_i = 1-I(Y_i > C_i)
\]

and $\phi(.)$ is a non-linear feature map. Furthermore, 
according to the Mercer's Theorem \cite{mercer1909}, 
$\mbold{k}(\mbold{X_i},\mbold{X_j})=
\phi(\mbold{X_i})^T\phi(\mbold{X_j})$,
as long as $\mbold{k(.,.)}$ is positive semi-definite. 
The $\xi_i$ and $\xi_i^\ast$ represent the slack variables.

The dual SSVM problem is then obtained as follows:
\[
\min_{\alpha,\alpha^{\ast}}
\frac{1}{2}\sum_{i=1}^{n}\sum_{k=1}^{n}(\alpha_i\alpha_k+\delta_i\delta_k\alpha_i^\ast\alpha_k^\ast)k(X_i,X_k)-\\
\sum_{i=1}^{n}\sum_{k=1}^{n}\alpha_i^\ast\delta_i\alpha_i k(X_i,X_k)-\sum_{i=1}^{n}\alpha_i Y_i+ \sum_{i=1}^{n}\delta_i \alpha_i^\ast Y_i 
\]
\[
\textrm{subject to} \left\{
\begin{matrix}
0  \leq \alpha_i\leq C  & \forall i \in {1,\dots,n} \\
0 \leq \alpha_i^\ast \leq C & \forall i \in {1,\dots,n} 
\end{matrix}\right.
\]

Dual SSVM problem is expressed in terms of the kernel $\mbold{k}(\mbold{X_i},\mbold{X_j})$, where $\alpha_i,\alpha_i^\ast$ are Lagrange multipliers. Consequently, the predictor obtained in the implicit non-linear feature space induced by $k(.,.)$ follows:
\begin{eqnarray}
h_{\text{SSVM}}(\mbold{X})&=&\sum_{i=1}^{n}(\alpha_i-\delta_i\alpha_i^{\ast})\mbold{k(X_i,X)}+b
\end{eqnarray}

\section{Concordance C-index for Survival\label{cindex}}

Consider the test set with survival outcomes $Y_{surv}=\{(Y_1,\delta_1),(Y_2,\delta_2),\ldots,(Y_n,\delta_n)\}$ and 
estimated prognostic indices $H_{surv}=\{h(\mbold{X_1}),h(\mbold{X_2}),\ldots,h(\mbold{X_n})\}$ obtained from the predictive survival model. 
Let the $T_i=(Y_i,\delta_i)$.
Then the C-index is calculated as follows \cite{vanBelleJMRL2011}:
\begin{eqnarray}
C &=& \frac{\sum_{i\neq j} \text{conc}(T_i,T_j,h(\mbold{X}_i),h(\mbold{X}_j))}{\sum_{i\neq j} \text{comp}(T_i,T_j)}
\end{eqnarray}

and
\begin{eqnarray}
\text{comp}(T_i,T_j)= \left\{
\begin{matrix} 1 &;& \text{if} \hspace{0.2cm} (Y_i<Y_j \hspace{0.1cm} \text{and} \hspace{0.1cm} \delta_i=1) \hspace{0.1cm} \text{or} \hspace{0.1cm} (Y_j<Y_i \hspace{0.1cm} \text{and} \hspace{0.1cm} \delta_j=1) \\
  0 &;& \hspace{-5.3cm}\text{otherwise}
  \end{matrix}\right.
\end{eqnarray}

\begin{eqnarray}
\text{conc}(T_i,T_j,h(\mbold{X}_i),h(\mbold{X}_j)) &=& \text{comp}(T_i,T_j) I[(h(\mbold{X}_j)-h(\mbold{X}_i))(Y_i-Y_j)>0]
\end{eqnarray}

\bibliography{wileyNJD-AMS}%

\begin{thebibliography}{41}
\expandafter\ifx\csname natexlab\endcsname\relax\def\natexlab#1{#1}\fi
\expandafter\ifx\csname url\endcsname\relax
  \def\url#1{{\tt #1}}\fi
\expandafter\ifx\csname urlprefix\endcsname\relax\def\urlprefix{URL }\fi
\expandafter\ifx\csname doiprefix\endcsname\relax\def\doiprefix{doi:}\fi

\bibitem[{Balcan et~al.(2008)Balcan, Blum, and N}]{balcan2008}
Balcan, M., A.~Blum, and S.~N, 2008: A theory of learning with similarity
  functions. {\it Machine Learning\/}, {\bf 72}, 89--112.

\bibitem[{Balog et~al.(2016)Balog, Lakshminarayanan, Ghahramani, Roy, and
  Teh}]{balog2016}
Balog, M., B.~Lakshminarayanan, Z.~Ghahramani, D.~M. Roy, and Y.~W. Teh, 2016:
  The mondrian kernel. {\it Proceedings of the 32nd Conference on Uncertainty
  in Artificial Intelligence\/}, 32–41.

\bibitem[{Biau and Scornet(2016)}]{biau2016}
Biau, G. and E.~Scornet, 2016: A random forest guided tour. {\it Test\/}, {\bf
  25}, 197--227.

\bibitem[{Bien and Tibshirani(2011)}]{bien2011}
Bien, J. and R.~Tibshirani, 2011: Prototype selection for interpretable
  classification. {\it Annals of Applied Statistics\/}, {\bf 5}, no. 4,
  2403--2424.

\bibitem[{Boulesteix et~al.(2012)Boulesteix, Janitza, Kruppa, and
  König}]{Boulesteix2012}
Boulesteix, A., S.~Janitza, J.~Kruppa, and I.~König, 2012: Overview of random
  forest methodology and practical guidance with emphasis on computational
  biology and bioinformatics. {\it Wire's Data Mining and Knowledge
  Discovery\/}, {\bf 2}, no. 6, 493--507.

\bibitem[{Breiman(2000)}]{breiman2000}
Breiman, L., 2000: Some infinity theory for predictor ensembles.  Technical
  Report 579, Statistics Dept. UCB.

\bibitem[{Chen et~al.(2009)Chen, Garcia, Gupta, et~al.}]{chen2009}
Chen, Y., E.~Garcia, M.~Gupta, et~al., 2009: Similarity-based classification:
  Concepts and algorithms. {\it Journal of Machine Learning Research\/}, {\bf
  10}, 747--776.

\bibitem[{Davies and Ghahramani(2014)}]{davies2014}
Davies, A. and Z.~Ghahramani, 2014: The random forest kernel and other kernels
  for big data from random partitions. {\it arXiv preprint arXiv:1402.4293\/}.

\bibitem[{Fedorov et~al.(2008)Fedorov, Mannino, and Zhang}]{fedorov2008}
Fedorov, V., F.~Mannino, and R.~Zhang, 2008: Consequences of dichotomization.
  {\it Pharmaceutical Statistics\/}, 50--61.

\bibitem[{Fernandez-Delgado et~al.(2014)Fernandez-Delgado, Cernadas, Barro, and
  Amorim}]{fernandezdelgado2014}
Fernandez-Delgado, M., E.~Cernadas, S.~Barro, and D.~Amorim, 2014: Do we need
  hundreds of classifiers to solve real world classification problems? {\it
  Journal of Machine Learning Research\/}, {\bf 15}, 3133--3181.

\bibitem[{Fouodo(2018)}]{survivalsvm}
Fouodo, C., 2018: Package 'survivalsvm'. {\it
  https://cran.r-project.org/web/packages/survivalsvm\/}.

\bibitem[{Friedman et~al.(2009)Friedman, Hastie, and
  Tibshirani}]{friedmanHastieTibshirani2009}
Friedman, J., T.~Hastie, and R.~Tibshirani, 2009: {\it The Elements of
  Statistical Learning\/}. Springer.

\bibitem[{Friedman(1991)}]{friedman}
Friedman, J.~H., 1991: Multivariate adaptive regression splines. {\it The
  annals of statistics\/}, 1--67.

\bibitem[{Harrell~Jr et~al.(1996)Harrell~Jr, Lee, and Mark}]{harrell1996}
Harrell~Jr, F.~E., K.~L. Lee, and D.~B. Mark, 1996: Multivariable prognostic
  models: issues in developing models, evaluating assumptions and adequacy, and
  measuring and reducing errors. {\it Statistics in medicine\/}, {\bf 15}, no.
  4, 361--387.

\bibitem[{Herbich(2001)}]{herbich2001}
Herbich, R., 2001: {\it Learning kernel classifiers\/}. MIT Press.

\bibitem[{Hothorn et~al.(2006)Hothorn, Hornik, and Zeileis}]{cforest}
Hothorn, T., K.~Hornik, and A.~Zeileis, 2006: Unbiased recursive partitioning:
  A conditional inference framework. {\it Journal of Computational and
  Graphical Statistics\/}, {\bf 15}, no. 3, 651--674.

\bibitem[{Ishwaran and Lu(2019)}]{ishwaran2019}
Ishwaran, H. and M.~Lu, 2019: Standard errors and confidence intervals for
  variable importance in random forest regression, classification, and
  survival. {\it Statistics in Medicine\/}, 558--582.

\bibitem[{James et~al.(2013)James, Witten, T, and R}]{James2013}
James, G., D.~Witten, H.~T, and T.~R, 2013: {\it An Introduction to Statistical
  Learning\/}. Springer.

\bibitem[{Kar and Jain(2011)}]{kar2011}
Kar, P. and P.~Jain, 2011: Similarity-based learning via data driven
  embeddings. {\it Pro- ceedings of the Advances in Neural Information
  Processing Systems,\/}, 1998–2006.

\bibitem[{Karatzoglou(2019)}]{kernlab}
Karatzoglou, A., 2019: Package 'kernlab'. {\it
  https://cran.r-project.org/web/packages/kernlab/kernlab.pdf\/}.

\bibitem[{Legendre and Legendre(2012)}]{legendre2012}
Legendre, P. and L.~F. Legendre, 2012: {\it Numerical ecology\/}. Elsevier.

\bibitem[{Linero(2017)}]{linero2017}
Linero, A.~R., 2017: A review of tree-based bayesian methods. {\it
  Communications for Statistical Applications and Methods\/}, {\bf 24}, no. 6,
  543--559.

\bibitem[{Marcus(2017)}]{marcus2017}
Marcus, R., 2017: The often-overlooked random forest kernels. {\it
  https://rmarcus.info/blog/2017/10/04/rfk.html\/}.

\bibitem[{Meier et~al.(2009)Meier, Van~de Geer, B{\"u}hlmann, et~al.}]{meier}
Meier, L., S.~Van~de Geer, P.~B{\"u}hlmann, et~al., 2009: High-dimensional
  additive modeling. {\it The Annals of Statistics\/}, {\bf 37}, no. 6B,
  3779--3821.

\bibitem[{Mercer(1909)}]{mercer1909}
Mercer, J., 1909: Functions of positive and negative type and their connection
  with the theory of integral equations. {\it Philosophical Transactions of the
  Royal Society\/}, 415--446.

\bibitem[{Olson and Wyner(2018)}]{olson2018}
Olson, M. and A.~Wyner, 2018: Consequences of dichotomization. {\it arXiv
  preprint arXiv:1812.05792\/}.

\bibitem[{Pace and Barry(1997)}]{pace1997}
Pace, K. and R.~Barry, 1997: Sparse spatial autoregressions. {\it Statistics \&
  Probability Letters\/}, {\bf 33}, no. 3, 291–297.

\bibitem[{Pekalska et~al.(2001)Pekalska, Paclik, and Duin}]{pekalska2001}
Pekalska, E., P.~Paclik, and R.~Duin, 2001: A generalized kernel approach to
  dissimilarity-based classification. {\it Journal of Machine Learning
  Research\/}, {\bf 2}, 175--211.

\bibitem[{{R Core Team}(2017)}]{Rcran}
{R Core Team}, 2017: {\it R: A Language and Environment for Statistical
  Computing\/}. R Foundation for Statistical Computing, Vienna, Austria.
\newline\urlprefix\url{https://www.R-project.org/}

\bibitem[{Rasmussen and Williams(2006)}]{rasmussen2006}
Rasmussen, C. and C.~Williams, 2006: {\it Gaussian Processes for Machine
  Learning\/}. MIT Press.

\bibitem[{Schoelkopf and Smola(2001)}]{schoelkopf2001}
Schoelkopf, B. and A.~Smola, 2001: {\it Learning with kernels\/}. MIT Press.

\bibitem[{Scornet(2016)}]{scornet2016}
Scornet, E., 2016: Random forests and kernel methods. {\it IEEE Transactions on
  Information Theory\/}, {\bf 62}, no. 3, 1485 -- 1500.

\bibitem[{Shivaswamy et~al.(2007)Shivaswamy, Chu, and M}]{shiwasvami2007}
Shivaswamy, P., W.~Chu, and J.~M, 2007: A support vector approach to censored
  targets. {\it Proceedings of the Seventh IEEE International Conference on
  Data Mining (ICDM)\/}, IEEE Computer Society, California, 655--660.

\bibitem[{Sokolov et~al.(2016)Sokolov, Carlin, Paul, et~al.}]{sokolov2016}
Sokolov, A., D.~Carlin, E.~Paul, et~al., 2016: Pathway-based genomics
  prediction using generalized elastic net. {\it PLOS Computational Biology\/},
  {\bf 12}, no. 3.

\bibitem[{van Belle et~al.(2011{\natexlab{a}})van Belle, Pelckmans, Van~Huffel,
  and Suykens}]{vanBelleJMRL2011}
van Belle, V., K.~Pelckmans, S.~Van~Huffel, and J.~Suykens, 2011{\natexlab{a}}:
  Learning transformation models for ranking and survival analysis. {\it
  Journal of Machine Learning Research\/}, {\bf 12}, 819--862.

\bibitem[{van Belle et~al.(2011{\natexlab{b}})van Belle, Pelckmans, Van~Huffel,
  and Suykens}]{vanBelle2011}
--- 2011{\natexlab{b}}: Support vector methods for survival analysis: a
  comparison between ranking and regression approaches. {\it Artificial
  Intelligence in Medicine\/}, {\bf 53}, 107--118.

\bibitem[{Van~der Laan et~al.(2007)Van~der Laan, Polley, and
  Hubbard}]{van2007super}
Van~der Laan, M.~J., E.~C. Polley, and A.~E. Hubbard, 2007: Super learner. {\it
  Statistical applications in genetics and molecular biology\/}, {\bf 6}, no.
  1.

\bibitem[{Wolpert(1996)}]{wolpert1996}
Wolpert, D., 1996: The lack of a priori distinctions between learning
  algorithms. {\it Neural Computation\/}, {\bf 9}, 1341 -- 1390.

\bibitem[{Wright and Ziegler(2017)}]{ranger}
Wright, M.~N. and A.~Ziegler, 2017: {ranger}: A fast implementation of random
  forests for high dimensional data in {C++} and {R}. {\it Journal of
  Statistical Software\/}, {\bf 77}, no. 1, 1--17, doi:10.18637/jss.v077.i01.

\bibitem[{Zafari et~al.(2019)Zafari, Zurita-Milla, and
  Izquierdo-Verdiguier}]{zafari2019}
Zafari, A., R.~Zurita-Milla, and E.~Izquierdo-Verdiguier, 2019: Evaluating the
  performance of a random forest kernel for land cover classification. {\it
  Remote Sensing\/}, {\bf 11}, no. 575.

\bibitem[{Zhu et~al.(2015)Zhu, Zeng, and Kosorok}]{zhu2015}
Zhu, R., D.~Zeng, and M.~R. Kosorok, 2015: Reinforcement learning trees. {\it
  Journal of the American Statistical Association\/}, {\bf 110}, no. 512,
  1770--1784.

\end{thebibliography}


\begin{thebibliography}{8}
\expandafter\ifx\csname natexlab\endcsname\relax\def\natexlab#1{#1}\fi
\expandafter\ifx\csname url\endcsname\relax
  \def\url#1{{\tt #1}}\fi
\expandafter\ifx\csname urlprefix\endcsname\relax\def\urlprefix{URL }\fi
\expandafter\ifx\csname doiprefix\endcsname\relax\def\doiprefix{doi:}\fi

\bibitem[{Allen(2011)}]{Allen2011}
Allen, T.~T., 2011: {\it Introduction to {D}iscrete {E}vent {S}imulation and
  {A}gent-based {M}odeling: {V}oting {S}ystems{,} {H}ealth {C}are{,}
  {M}ilitary{,} and {M}anufacturing\/}. Springer, New York.

\bibitem[{Ballen(2011)}]{Ballen2011}
Ballen, T.~T., 2011: {\it Introduction to {D}iscrete {E}vent {S}imulation and
  {A}gent-based {M}odeling: {V}oting {S}ystems{,} {H}ealth {C}are{,}
  {M}ilitary{,} and {M}anufacturing\/}. 2nd ed., Springer, New York.

\bibitem[{Blanchard and Loubere(2015)}]{Blanchard2015}
Blanchard, G. and R.~Loubere, 2015: {\it High-order {C}onservative {R}emapping
  with a posteriori {MOOD} stabilization on polygonal meshes\/}. Available
  from: \url{http://www.emn.fr/z-info/choco-solver/} [last accessed {M}ay
  2011].

\bibitem[{Elbaum et~al.(February 2002)Elbaum, Malishevsky, and
  Rothermel}]{Elbaum2002}
Elbaum, S., A.~G. Malishevsky, and G.~Rothermel, February 2002: Test case
  prioritization: a family of empirical studies. {\it IEEE {T}ransactions on
  {S}oftware {E}ngineering\/}, {\bf 28}, no. 2, 159--182, doi:12345.2345.

\bibitem[{Rothermel(1997)}]{Rothermel1997}
Rothermel, G., 1997: A safe efficient regression test selection technique. {\it
  ACM {T}ransactions on {S}oftware {E}ngineering and {M}ethodology\/}, {\bf 6},
  no. 2, 173--210.

\bibitem[{Rothermel et~al.(1998)Rothermel, Harrold, Hirt, Amsden, and
  Cook}]{Rothermel1998}
Rothermel, G., M.~J. Harrold, C.~W. Hirt, A.~A. Amsden, and J.~L. Cook, 1998: A
  safe efficient regression test selection technique. {\it {ACM} {T}ransactions
  on {S}oftware {E}ngineering and {M}ethodology\/}, {\bf 6}, no. 2, 173--210.

\bibitem[{Schulz and Doblhammer(2012)}]{Schulz2012}
Schulz, A. and G.~Doblhammer, 2012: Aktueller und zuk{\"{u}}nftiger
  {K}rankenbestand von {D}emenz in {D}eutschland auf basis der outinedaten der
  {AOK}. ({C}urrent and future number of people suffering from dementia in
  {G}ermany based on routine data from the {AOK}.). {\it Versorgungs-Report\/},
  C.~Gnster, J.~Klose, and N.~Schmacke, Eds., IEEE {P}ress, Piscataway, NJ,
  USA, 161--175.

\bibitem[{Yoo and Harman(2007)}]{Yoo2007}
Yoo, S. and M.~Harman, 2007: Pareto efficient multi-objective test case
  selection. {\it Proceedings of the {I}nternational {C}onference on {S}oftware
  {T}esting and {A}nalysis\/}, London, UK, 140--150.

\end{thebibliography}

\section*{Author Biography}

Dai Feng holds a Ph.D. from the University of Iowa, and is currently employed by AbbVie Inc. Richard Baumgartner holds a Ph.D. from Technische Universit\"{a}t Wien, and is currently employed by Merck \& Co., Inc.

\end{document}